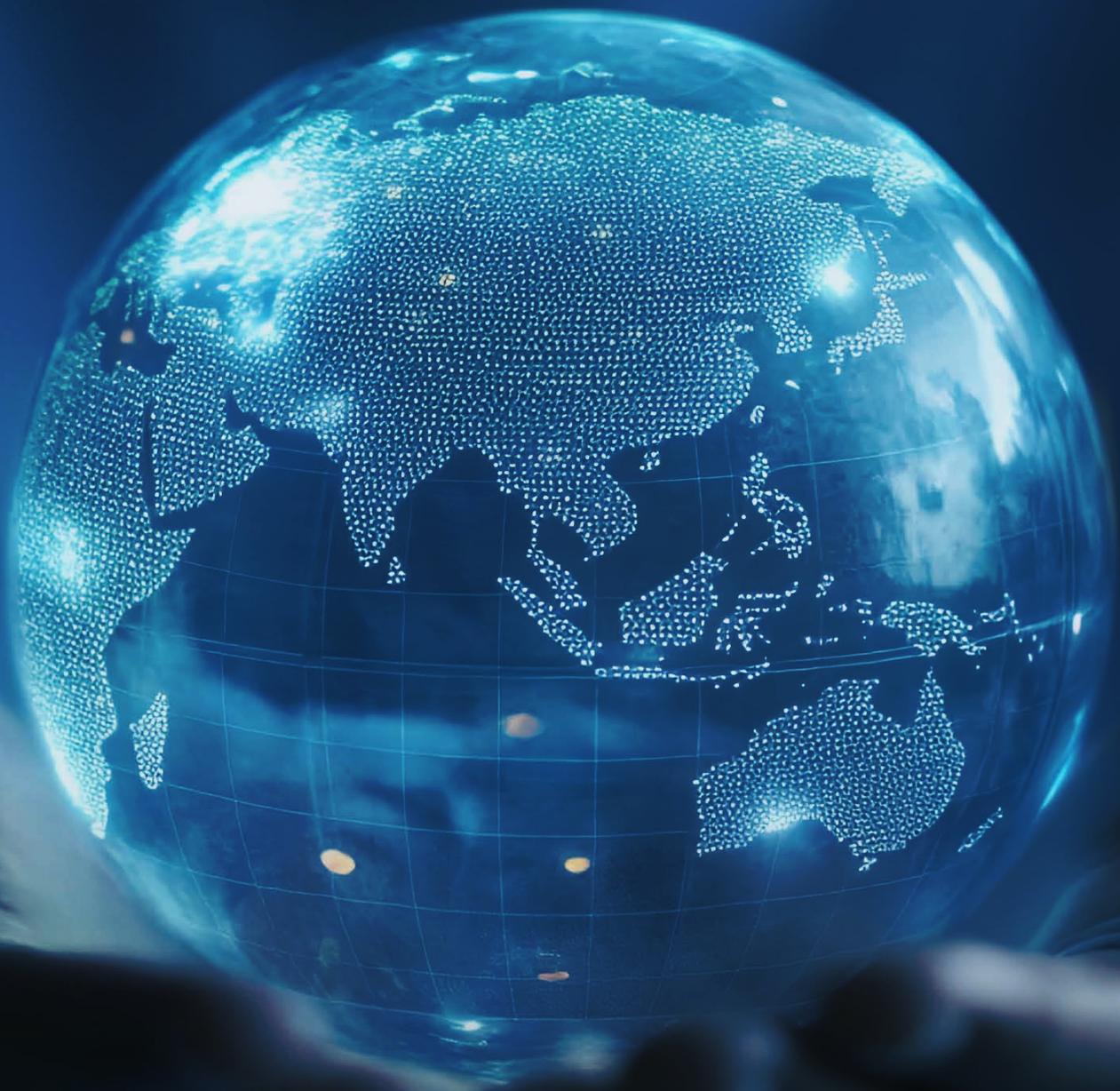

# The Singapore Consensus on Global AI Safety Research Priorities

Building a Trustworthy, Reliable and Secure AI Ecosystem

8 May 2025

TABLE OF CONTENTS

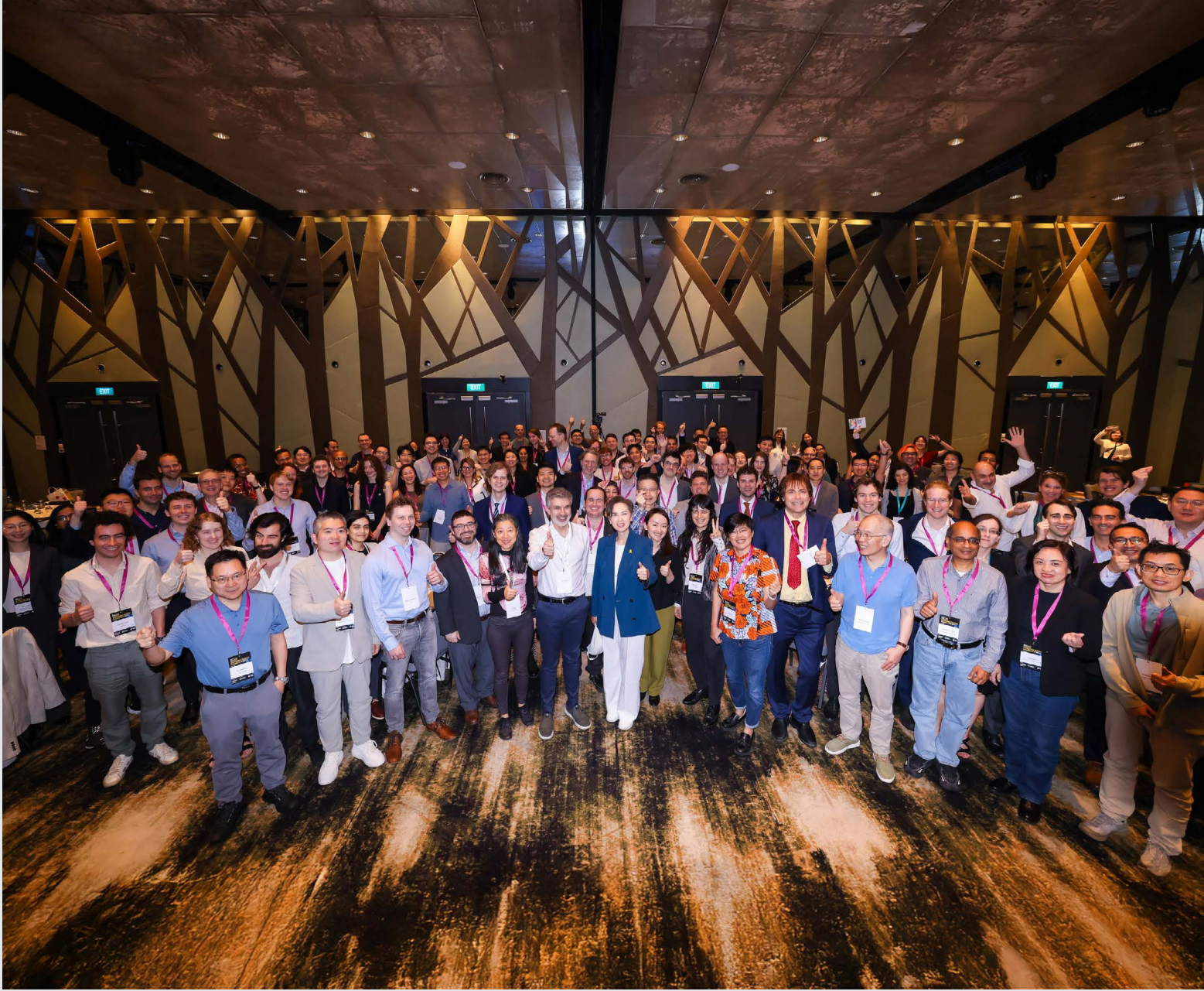

# Contributors to the Singapore Consensus

**Expert Planning Committee:**

**Dawn Song** | UC Berkeley

**Lan Xue** | Tsinghua University

**Luke Ong** | Nanyang Technological University

**Max Tegmark** | MIT

**Stuart Russell** | UC Berkeley

**Tegan Maharaj** | MILA

**Ya-Qin Zhang** | Tsinghua University

**Yoshua Bengio** | MILA

**Writing Coordinators:**

**Max Tegmark** | MIT

**Sören Mindermann** | MILA – Quebec Artificial Intelligence Institute / International AI Safety Report

**Stephen Casper** | MIT

**Vanessa Wilfred** | Infocomm Media Development Authority

**Wan Sie Lee** | Infocomm Media Development Authority

*The views expressed by contributors are in their individual capacities and do not necessarily reflect those of their affiliated organisations.*

**Image:** Participants of the '2025 *Singapore Conference on AI: International Scientific Exchange on AI Safety*', 26th April 2025.

## Individual Contributors:

**Abhishek Mishra** | Faculty AI
**Adam Gleave** | FarAI
**Adam Kalai** | Open AI
**Agnes Delaborde** | France INESIA (LNE)
**Akiko Murakami** | Japan AISI
**Alejandro Ortega** | Apollo Research
**Andrea Phua** | Ministry of Digital Development and Information
**Anthony Tung** | National University of Singapore
**Anu Sinha** | Google DeepMind
**Benjamin Prud'homme** | MILA - Quebec Artificial Intelligence Institute
**Besmira Nushi** | Microsoft Research
**Brian Tse** | ConcordiaAI
**Bryan Low** | National University of Singapore
**Buck Shlegeris** | Redwood Research
**Chaochao Lu** | Shanghai AI Laboratory
**Chen Hui Ong** | Infocomm Media Development Authority
**Cheston Tan** | A*STAR
**Chris Meserole** | Frontier Model Forum
**Chua Kuan Seah** | Cyber Security Agency of Singapore
**Cyrus Hodes** | 1infinity Ventures
**Dan Hendrycks** | xAI
**David Dalrymple** | Advanced Research + Invention Agency
**Denise Wong** | Infocomm Media Development Authority
**Dennis Duan** | Google DeepMind
**Djordje Zikelic** | Singapore Management University
**Duncan Cass-Beggs** | Global AI Risk Initiative
**Elizabeth Hilbert** | Meta
**Erik Jones** | Anthropic
**Fazl Barez** | University of Oxford
**Francisco Eiras** | Dynamo AI
**Freya Hempleman** | UK AISI

**Fynn Heide** | Safe AI Forum
**Hiromu Kitamura** | Japan AISI
**Hongjiang Zhang** | Co-Founder and Chairman WIZ.AI, and Chief Advisor to One North Foundation
**Ima Bello** | Future of Life Institute
**Jacob Steinhardt** | Transluce
**James Petrie** | University of Oxford & Future of Life Institute
**Jeff Alstott** | RAND
**Jeff Clune** | University of British Columbia / CIFAR AI
**Jeffrey Ladish** | Palisade Research
**Jiahao Chen** | New York City Office of Technology and Innovation
**Jiajun Liu** | CSIRO
**Jingren Wang** | Shanghai Artificial Intelligence Lab
**Jingyu Wang** | Tsinghua University (I-AIIG)
**Jinyu Fan** | Chinese Academy of Sciences
**Johannes Heidecke** | OpenAI
**John Willes** | Vector Institute
**Jong Lee** | Google DeepMind
**Joshua Engels** | Massachusetts Institute of Technology
**Julian Michael** | Scale AI
**Juntao Dai** | Beijing Academy of Artificial Intelligence
**Karine Perset** | Organisation for Economic Co-operation and Development
**Kihyuk Nam** | Korea AISI
**Kwan Yee Ng** | ConcordiaAI
**Kwok Yan Lam** | Singapore AISI
**Lily Sun** | Massachusetts Institute of Technology
**Malo Bourgon** | Machine Intelligence Research Institute
**Mark Brakel** | Future of Life Institute
**Mark Nitzberg** | Center for Human-Compatible AI
**Martin Soto** | UK AISI
**Mathias Kirk Bonde** | ControlAI

**Max Tegmark** | Massachusetts Institute of Technology
**Meindert Kamphuis** | Ministry of Economic Affairs and Climate
**Michael Belinsky** | Schmidt Sciences
**Miguel Fernandes** | Resaro
**Minlie Huang** | Tsinghua University
**Miro Plueckebaum** | Singapore AI Safety Hub / Oxford Martin AIGI
**Mohan Kankanhalli** | National University of Singapore
**Nayat Sanchez-Pi** | Inria / Inria Chile / Binational Center on AI (France-Chile)
**Nicolas Moes** | The Future Society
**Nitarshan Rajkumar** | University of Cambridge
**Noah Goodman** | Google DeepMind
**Nouha Dziri** | Allen Institute
**Pierre Peigné** | PRISM Eval
**Reihaneh Rabbany** | McGill University and MILA - Quebec Artificial Intelligence Institute
**Richard Mallah** | Center for AI Risk Management & Alignment
**Robert Trager** | Oxford Martin AI Governance Initiative
**Rongwu Xu** | Tsinghua University
**Ruimin He** | Ministry of Digital Development and Information
**Rumman Chowdhury** | Humane Intelligence
**Saad Siddiqui** | Safe AI Forum
**Sami Jawhar** | Model Evaluation & Threat Research
**Sarah Schwettmann** | Transluce
**Seán Ó hÉigeartaigh** | Center for the Study Existential Risk
**Shameek Kundu** | AI Verify Foundation
**Shao Wei Ying** | NCS Pte Ltd

**Shayne Longpre** | Massachusetts Institute of Technology
**Siméon Campos** | SaferAI
**Simon Moeller** | EU AI Office
**Sören Mindermann** | MILA - Quebec Artificial Intelligence Institute / International AI Safety Report
**Stephen Casper** | Massachusetts Institute of Technology
**Subhash Kantamneni** | Massachusetts Institute of Technology
**Taewhi Lee** | Korea AISI
**Tammy Masterson** | UK AISI
**Thomas Kwa** | Model Evaluation & Threat Research
**Ting Dong** | Tsinghua University
**Vanessa Wilfred** | Infocomm Media Development Authority
**Vidhisha Balachandran** | Microsoft Research
**Wan Sie Lee** | Infocomm Media Development Authority
**Wei Xu** | Tsinghua University
**William Tjhi** | AI Singapore
**Xiaojian Li** | Tsinghua University
**Yi Zeng** | Chinese Academy of Sciences
**Yifan Mai** | Stanford University
**Yong Kiam Tan** | Nanyang Technological University
**Zico Kolter** | Carnegie Mellon University

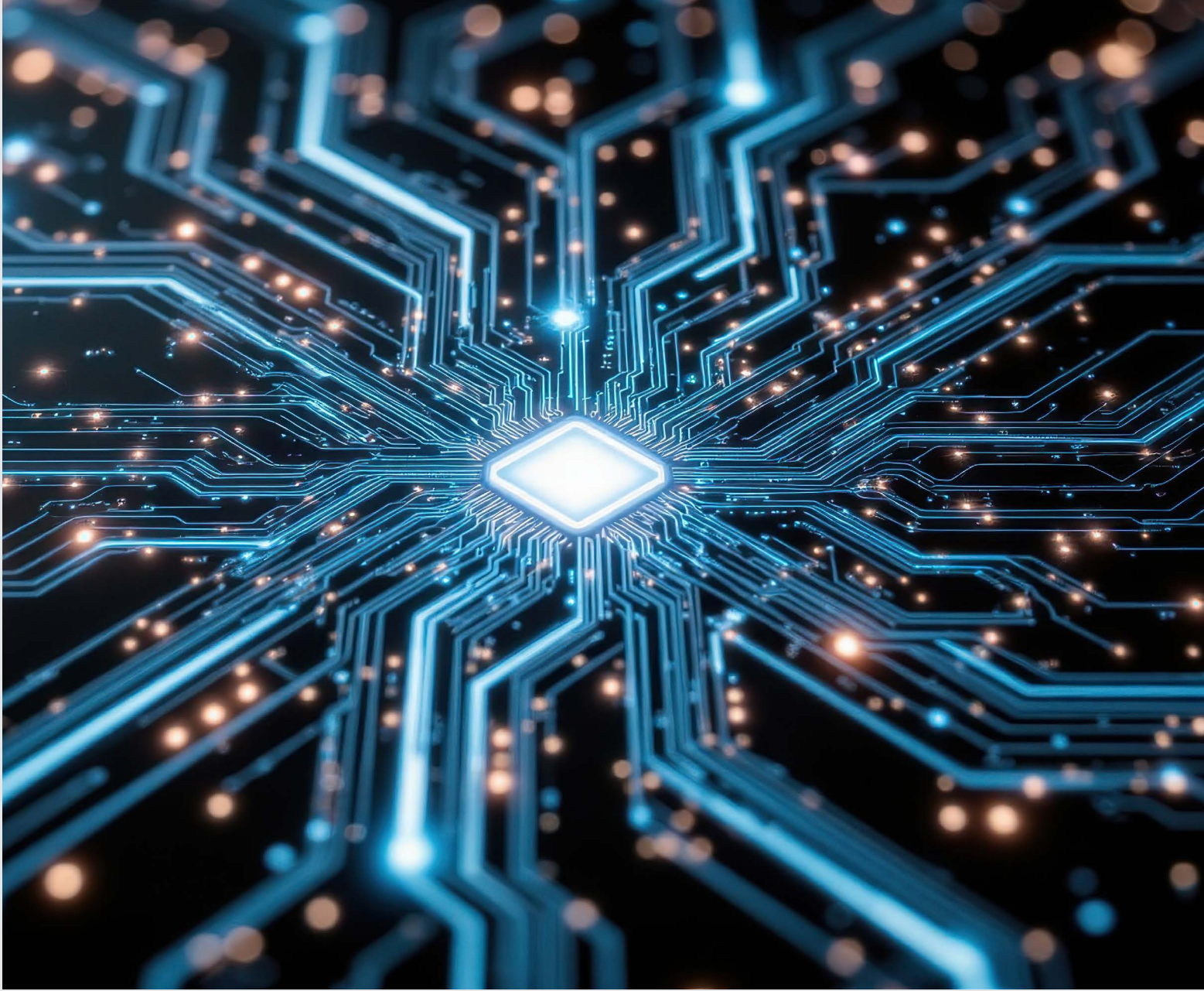

# Introduction

### Building a Trusted Ecosystem

 Rapidly improving AI capabilities and autonomy hold significant promise of transformation, but are also driving vigorous debate on how to ensure that AI is safe, i.e., trustworthy, reliable, and secure. Building a trusted ecosystem is therefore essential – it helps people **embrace AI with confidence** and gives **maximal space for innovation** while avoiding backlash. This requires policymakers, industry, researchers and the broader public to collectively work toward securing positive outcomes from AI's development. **AI safety research is a key dimension**. Given that the state of science today for building trustworthy AI does not fully cover all risks, accelerated investment in research is required to keep pace with commercially driven growth in system capabilities.

## Goals

The *2025 Singapore Conference on AI (SCAI): International Scientific Exchange on AI Safety* aims to support research in this important space by bringing together AI scientists across geographies to **identify and synthesise research priorities in AI safety**. The result, *The Singapore Consensus on Global AI Safety Research Priorities,* builds on the [International AI Safety Report-A](#) (IAISR) chaired by Yoshua Bengio and backed by 33 governments. By adopting a defence-in-depth model, this document organises AI safety research domains into three types: challenges with creating trustworthy AI systems (Development), challenges with evaluating their risks (Assessment), and challenges with monitoring and intervening after deployment (Control).

Through the Singapore Consensus, we hope to **globally facilitate meaningful conversations between AI scientists and AI policymakers for maximally beneficial outcomes**. Our goal is to enable more impactful R&D efforts to rapidly develop safety and evaluation mechanisms and foster a trusted ecosystem where AI is harnessed for the public good.

**Areas of mutual interest:** While companies and nations often compete on AI research and development, there are also incentives to find alignment and common interests. This synthesis covers areas where different parties may be competitive, but also highlights examples from the broader landscape of areas of mutual interest – research products and information that developers would find it in their self-interest to share widely ([Bucknall-B](#)). Certain safety advances offer minimal competitive edge while serving a common interest – similar to how competing aircraft manufacturers (e.g., Boeing and Airbus) collaborate on aviation safety information and standards. In AI, particular areas for potentially mutually-beneficial cooperation span sections 1-3 of this report and include certain verification mechanisms, risk-management standards, and risk evaluations ([Bucknall-B](#)). The motivation is clear: no organisation or country benefits when AI incidents occur or malicious actors are enabled, as the resulting harm would damage everyone collectively.

## Process

| Key event | Contributors | Representation |
|---|---|---|
| 26th April 2025 – SCAI: International Scientific Exchange on AI Safety | More than 100 participants in attendance for discussion and feedback | Participants from 11 countries were present |

This document represents a comprehensive synthesis of research proposals drawn from the [International AI Safety Report-B](#) and complementary recent research prioritisation frameworks, including [UK AISI](#), [Anthropic-F](#), [Anwar](#), [Bengio-A](#), [GDM](#), [Hendrycks-A](#), [Ji](#), [Li-A](#), [OpenAI-B](#), [NIST](#), [Reuel](#), [Slattery](#), and [Weidinger-A](#). Initially designed as a consultation draft by the **Expert Planning Committee** (Dawn Song, Lan Xue, Luke Ong, Max Tegmark, Stuart Russell, Tegan Maharaj, Ya-Qin Zhang, and Yoshua Bengio), it was distributed to all conference participants to solicit comprehensive feedback. Following several rounds of updates based on further participant feedback in writing and in person, this document has been designed to synthesise points of broad consensus among diverse researchers. The full list of **conference participants** who contributed to this Singapore Consensus process is presented at the beginning of this document, and includes researchers from leading academic institutions and AI developers, as well as representatives from governments and civil society.

We have attempted to be inclusive of both terminology and research topic suggestions from researchers in academia, industry, and civil society. This synthesis presented unique challenges because different authors have used a variety of non-equivalent definitions and classification schemes. This report therefore takes a humble approach: the definitions of key terms in Table 1 below simply specify how we use various terms in this report, to avoid confusion, and we make no claims whatsoever to these being better than other alternative definitions.

## Scope

We limit our discussion to technical AI safety research, focused on making AI more trustworthy rather than merely more powerful, and excluding AI policy research. We focus primarily on **general-purpose AI**: Following the International AI Safety Report, the term 'AI systems' in this document should be understood to refer to general-purpose AI (GPAI) systems – systems that can perform or can be adapted to perform a wide range of tasks (IAISR). This includes language models that produce text (e.g. chat systems) as well as 'multimodal' models which can work with multiple types of data, often including text, images, video, audio, and robotic actions. Importantly, it includes general-purpose agents – systems that autonomously act and plan to accomplish complex tasks, for example by controlling computers. Developing more powerful agents is a core focus of AI developers as their growing deployment presents new major risks and opportunities.

We emphasise that *technical* solutions relating to *general-purpose AI systems* are **necessary** but **not sufficient** for the overall safety of AI. Our collective ability to responsibly manage AI risks and opportunities will ultimately depend on our choices to build a healthy AI ecosystem, study risks, implement mitigations, and integrate solutions into effective risk management frameworks.

## Structure

Inspired by the 2025 International AI Safety Report (IAISR), this document adopts a defence-in-depth model and groups technical AI safety research topics into three broad areas from **risk assessment** that informs subsequent development and deployment decisions, to technical methods in the system **development** phase, and tools for **control** after a system has been deployed. The three identified areas have interesting overlaps as illustrated in Figure 1:

1. **Risk Assessment**: The primary goal of risk assessment is to understand the severity and likelihood of a potential harm. Risk assessments are used to prioritise risks and determine if they cross thresholds that demand specific action. Consequential development and deployment decisions are predicated on these assessments. The research areas in this category involve developing methods to measure the **impact of AI systems** for both current and future AI, enhancing **metrology** to ensure that these measurements are precise and repeatable, and building enablers for **third-party**

   **audits** to support independent validation of these risk assessments.

2. **Development**: AI systems that are trustworthy, reliable and secure by design give people the confidence to embrace and adopt AI innovation. Following the classic safety engineering framework, the research areas in this category involves **specifying** the desired behaviour, **designing** an AI system that meets the specification, and **verifying** that the system meets its specification.

3. **Control**: In engineering, "control" usually refers to the process of managing a system's behaviour to achieve a desired outcome, even when faced with disturbances or uncertainties, and often in a feedback loop. The research areas in this category involve developing monitoring and intervention mechanisms for **AI systems**, extending monitoring mechanisms to the broader **AI ecosystem** to which the AI system belongs, and societal resilience research to strengthen **societal infrastructure** (e.g. economic, security) to adapt to AI-related societal changes.

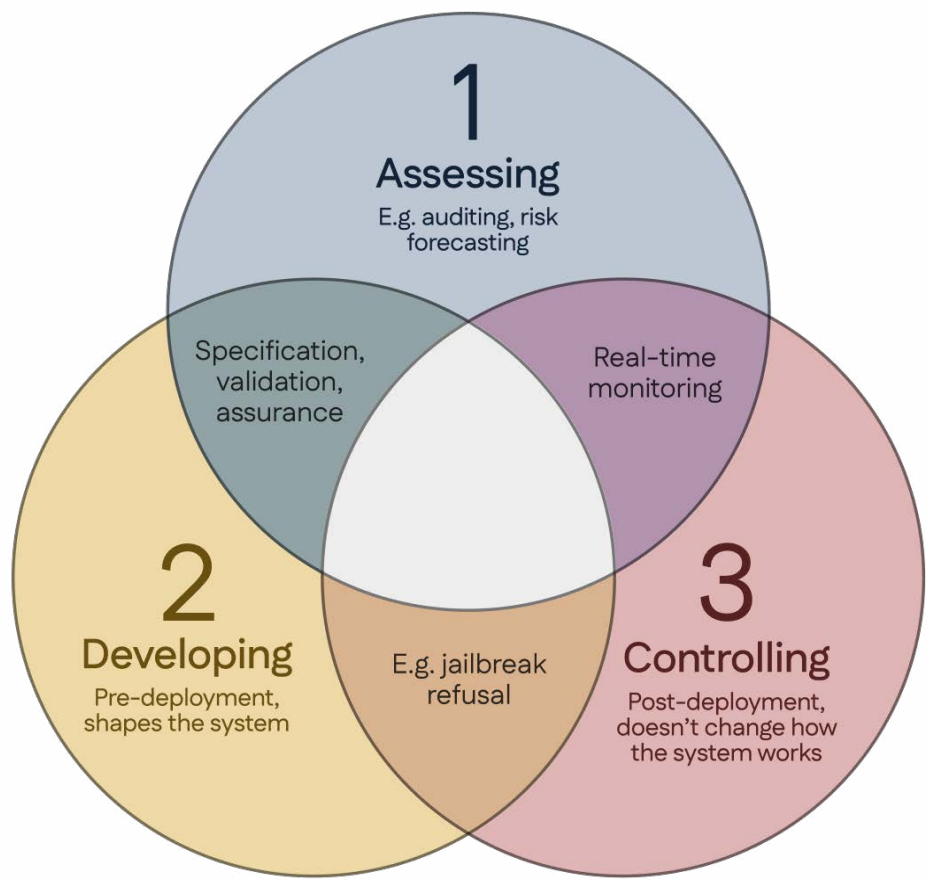

**Figure 1:** This venn diagram illustrates how AI safety techniques are related. We make a system behave as desired (assessed in Area 1) both by how we design it (Area 2) and how we control it afterward (Area 3). Some assessment tools can support both Areas 2 and 3. Overlap between Areas 2 and 3 stems from different choices for what to count as part of the system versus controlling feedback loops and scaffolds. For example, an external filter that blocks questions about bioweapons falls into Area 2 if we consider the large language model to be the system, but in Area 1 if we consider the filter to be part of the system.

| Term | How it is used in this report |
| --- | --- |
| Specification | Specific definition of desired system behaviour. |
| Validation | Ensuring that the specification and the final system meets the needs of the user, developer, or society (did I build the right system?) . |
| Validity | How well a measurement or assessment tool actually measures what it claims to measure. |
| AI agent | An AI which can make plans to achieve goals, adaptively perform tasks involving multiple steps and uncertain outcomes along the way, and interact with its environment – for example by creating files, taking actions on the web, or delegating tasks to other agents – with little to no human oversight. |
| AI model | A computer program, often automatically created by learning from data, designed to process inputs and generate outputs. AI models can perform tasks such as prediction, classification, decision-making, or generation, forming the engine of AI systems. |
| AI system | An integrated setup that combines one or more AI models with other components, such as user interfaces or content filters, to produce an application that users can interact with. |
| Verification | Providing qualitative or quantitative justifications or guarantees that a system meets its specification (did I build the system right?). |
| Assurance | The broader process of determining if a system performs as intended. As such, providing assurance requires appropriate specification, validation, design, implementation and verification. |
| Control | Monitoring a system after it has been created and intervening where needed, often in a feedback loop, to ensure the system behaves as desired. |
| Alignment | Creating/modifying AI to meet intended behaviour, goals, and values (current emphasis tends to be on behaviour). |
| Intelligence | Ability to accomplish goals. |
| Artificial intelligence (AI) | Non-biological intelligence. |
| Narrow intelligence | Ability to accomplish goals in a narrow domain, e.g. chess. |
| Artificial general intelligence (AGI) | AI that can do most cognitive work as well as humans. This implies that it is highly autonomous and can do most economically valuable remote work as well as humans. |
| Artificial super-intelligence (ASI) | AI that can accomplish any cognitive work far beyond human level. |

**Table 1:** Glossary of how we use key terminology in this report. Specification, validation, assurance, and verification are central concepts in systems engineering. **Note:** *Different authors have used a variety of non-equivalent definitions. The definitions in this table simply specify how this report uses various terms, not how they should be used in general. We use the terms "AGI", "ASI" and "intelligence" much as in the original definitions by Gubrud, Legg, and Bostrom.*

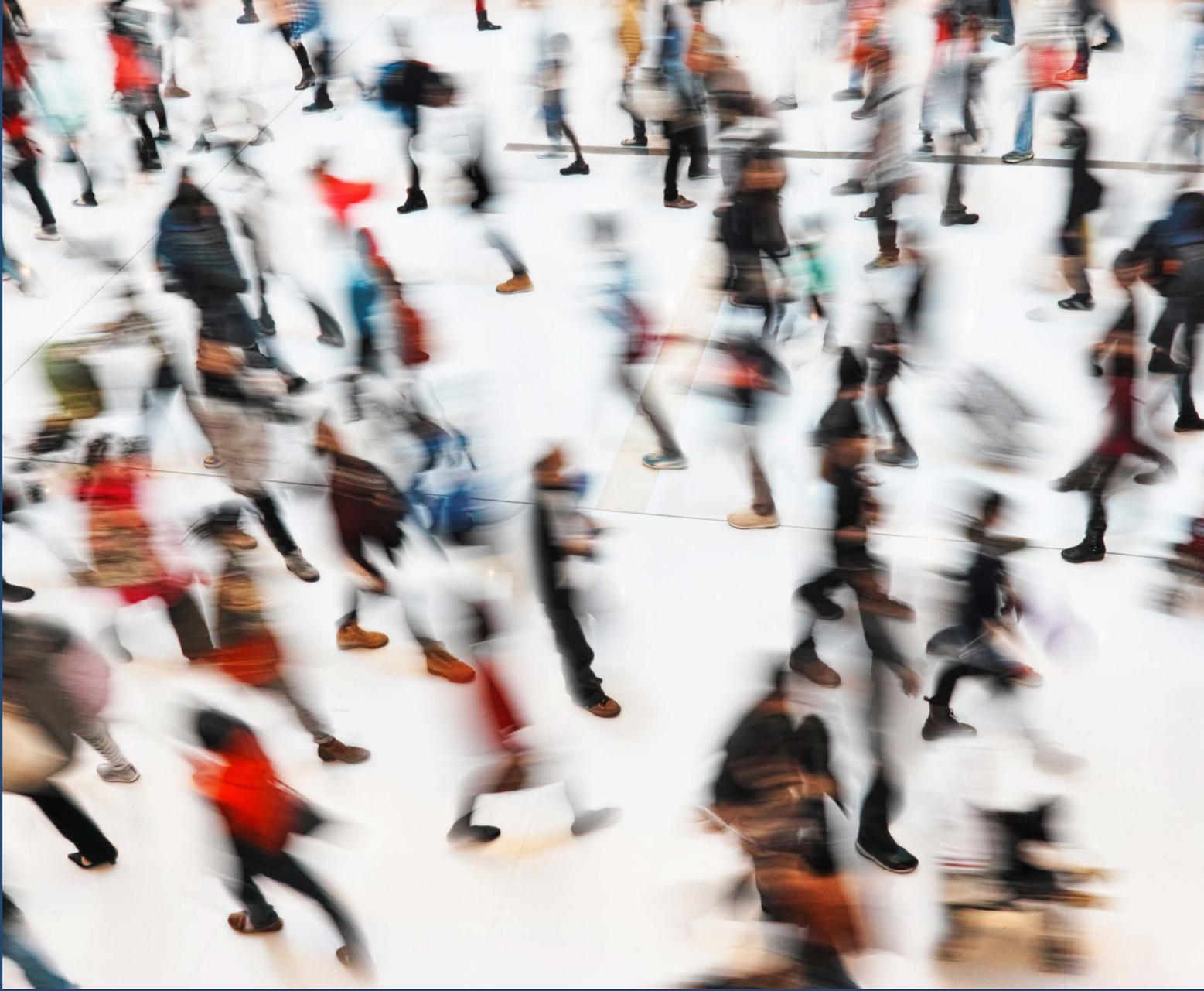

# 1 Risk Assessment

Associated with IAISR chapter 3.3

The primary goal of risk assessment is to understand the severity and likelihood of a potential harm. Risk assessments are used to prioritise risks and determine if they cross thresholds that demand specific action. Consequential development and deployment decisions are predicated on these assessments. The research areas in this category involve:

A. *Developing methods to measure the impact of AI systems for both current and future AI* – This includes developing standardised assessments for **risky behaviours of AI systems** through audit techniques and benchmarks, evaluation and assessment of new capabilities, including potentially dangerous ones; and for **real-world societal impact** such as labour, misinformation and privacy through field tests and prospective risks analysis.

B. *Enhancing metrology to ensure that the measurements are precise and repeatable* – This includes research in technical methods for **quantitative risk assessment** tailored to AI systems to reduce uncertainty and the need for large safety margins. This is an important open area of research.

> C. *Building enablers for third-party audits to support independent validation of risk assessments* – This includes developing **secure infrastructure** that enables thorough evaluation while protecting intellectual property, including preventing model theft.

Existing AI regulations and AI company commitments require rigorous risk identification and assessment, and consequential deployment decisions are predicated on these assessments (e.g. EU, OpenAI-A, Anthropic-E, Google). The primary goal of risk assessment is to understand the severity and likelihood of a potential harm. Risk assessments are used to *prioritise* risks and to determine if they cross *risk thresholds* that demand specific action such as mitigation. These thresholds – often defined in terms of measurable key risk indicators such as model evaluations (Campos) – are useful markers for further intervention. For example, if a system is found to have the ability to substantially assist malicious users in conducting cyberattacks, this may be considered an unacceptable risk. Risk assessment also informs safer development practices (Section 2) and control practices (Section 3) needed to mitigate risks.

Carefully defined risk thresholds are this report's first example of a potential **area of mutual interest** – actors may find it in their self-interest to share them widely or cooperate on them, even with competitors. This report highlights several further examples of these areas, but does not highlight every example explicitly.

The research areas in this category involve developing methods to study the **present impacts** of AI systems and forecast their potential **future implications**.

## 1.1 Audit techniques and benchmarks

Techniques and benchmarks with which AI systems can be effectively and efficiently tested for harmful behaviours are highly varied and central to risk assessments (IAISR, Birhane-A). However, developing high-quality standardised assessments for AI system capabilities is difficult due to the inability of research benchmarks to capture complexities in the real world (Raji, Eriksson). For example, different evaluations of AI systems' values can produce very different results depending on simple aspects of experimental design (Khan). Frontiers for future work include the generation of high-quality evaluations, dynamic automated evaluations, developing technical "red lines" or risk thresholds, establishing best practices for translating research findings into evaluation protocols that can be standardised across different organisations, and incorporating auditing into decision-making frameworks. It will also be key to keep evaluation resources secure and maintain them over time to reduce the possibility of developers gaming them.

## 1.2 Downstream impact assessment and forecasting

Assessing and forecasting the many societal impacts of AI systems is one of the most central goals of risk assessments. However, it is also very challenging due to its inherent prospective and complex nature (Weidinger-B, Solaiman). Research on forecasting involves studying usage data, analysing trends, risk modelling, predicting progress in AI capabilities, developing models of AI's future impacts, and updating forecasts in response to findings from field tests and usage data. This research also plays an important role in informing which evaluations and audits are needed for *valid* assessments of likely and severe risk scenarios. Because of the complexities involved in the study of downstream societal impacts, continued work to thoroughly monitor and study them will require nuanced analysis, interdisciplinarity, and inclusion (Wallach).

### 1.2.1 Field tests

Field tests and human participant studies aim to assess the real-world impacts of AI systems. They include analysing current impacts on topics such as deepfakes, labour, inequality, market concentration, misinformation, polarisation, privacy, mental health, and education. For example, some researchers have published details of how AI systems are being used and which professions they are affecting, aiming to inform decision-makers on the economic and labour market impacts of AI (Anthropic-A). Developers also sometimes start "bug bounty programs" to incentivise users to find and report vulnerabilities so that they can be fixed (e.g. Anthropic-H). One kind of field test that is particularly relevant to malicious use risks is "uplift studies" (Bateman). Uplift studies aim to assess how much an AI system can help users with a task (e.g. performing cyberattacks) relative to users without access to that system. For example, some AI labs have tested if using LLMs uplifts humans' abilities to plan biological attacks (OpenAI-C). Field tests, combined with other usage data can also assess questions such as how an AI system affects the mental health of users. As in the field of clinical drug trials, field tests may start with limited, controlled tests, and then gradually expand to real-world contexts to uncover new risks and side effects.

### 1.2.2 Prospective risk analysis and structured analytical techniques

The International AI Safety Report (IAISR) highlights the 'evidence dilemma' for emerging AI risks. On the one hand, early mitigations for emerging risks can turn out to be unnecessary or ineffective. On the other hand, waiting for clear evidence of a risk before mitigating it can leave society unprepared or even make mitigation impossible.

To navigate this dilemma, transparency infrastructure and early risk assessment are key. When assessing risks that have not yet occurred, or risks that may take a variety of forms (e.g. cyber attacks), it is often necessary to use prospective risk analysis and structured analytical techniques. These techniques are often used outside the field of AI, e.g. in nuclear safety, cybersecurity, or aircraft flight control. They have also been crucial in historical debates, e.g. over the health impacts of ozone depletion and smoking. Nonetheless, they are not yet widely used in AI risk assessment (IAISR, Murray, Casper-C).

Prospective risk analysis and structured analytical techniques include (IAISR):
- Explorative foresight: Scenario analysis and planning; horizon scanning; threat modelling and risk modelling.
- Probabilistic risk assessment (often used in high-reliability industries like nuclear and aerospace).
- Judgement elicitation and integration (e.g. Delphi method).
- Systems thinking.
- Causal mapping techniques (e.g. Bow-Tie analysis, Event Tree Analysis).

Structured risk assessments are also needed to combine evidence in order to construct a *safety case*, used by an AI developer to convincingly demonstrate that their system is safe (Clymer, Buhl). This requires assessing the full life cycle and the full stack of safety techniques used, as well as an assessment of the systemic interaction between components and the outside world (see 1.6).

## 1.3 Secure evaluation infrastructure

External auditors and oversight bodies need infrastructure and protocols that enable thorough evaluation while protecting sensitive intellectual property. Ideally, evaluation infrastructure should enable double-blindness: the evaluator's inability to directly access the system's parameters and developers' inability to know what exact evaluations are run (Reuel, Bucknall-A, Casper-B). Meanwhile, the importance of mutual security will continue to grow as system capabilities and risks increase. Methods for developing secure infrastructure for auditing and oversight are known to be possible. However, open challenges include determining what level of access is appropriate for which evaluations and conducting the engineering work of designing, building, and integrating efficient infrastructure. Further research should also explore how audit results can be effectively and reliably incorporated into risk management and decision-making frameworks.

## 1.4 System safety assessment

Safety assessment is not just about individual AI systems, but also their interaction with the rest of the world. For example, when an AI company discovers concerning behaviour from their system, the resulting risks depend, in part, on having internal processes in place to escalate the issue to senior leadership and work to mitigate the risks.

System safety considers both AI systems and the broader context that they are deployed in. The study of system safety focuses on the interactions between different technical components as well as processes and incentives in an organisation (IAISR, Hendrycks-B, AISES, Alaga). The practice of system safety engineering has a long history in areas such as aircraft flight control and nuclear reactor control (Dekker). System safety assessments evaluate if a critical system continues to function as intended even under human error, insider threats, or the failure of individual technical components. In AI safety assessments, this includes analysing how AI deployments might interact with existing social, economic, and political structures to create emergent downstream risks that individual system evaluations might miss (Weidinger-B), as well as analysing risks that emerge from multiple AI systems and humans interacting with each other.

## 1.5 Metrology for AI risk assessment

Metrology, the science of measurement, has only recently been studied in the context of AI risk assessment (IAISR, Hobbhahn). Current approaches generally lack standardisation, repeatability, and precision. For example, existing measurements such as benchmarks and audits often exhibit weak internal validity (ensuring assessments measure actual capabilities rather than test-taking artifacts), external validity (addressing how well test results generalise to real-world deployment contexts), and construct validity (accurately measuring abstract safety-relevant concepts such as deception or power-seeking tendencies). Typical approaches to quantitative risk assessment come from the field of actuarial risk assessment, (i.e. the insurance industry). While these risk assessment methods can be very useful for quantifying and studying risks which are easily associated with monetary damages, they can also fail to capture other kinds of risk, for example those which arise from interaction of multiple risks, or from systemic factors not easily quantified. Research in technical methods for quantitative risk assessment tailored to AI systems is an important open area. Enhanced metrology would reduce uncertainty and the need for large safety margins, enabling more reliable comparisons across AI systems and more precise identification of KRIs such capability thresholds that trigger risk thresholds.

## 1.6 Dangerous capability and propensity assessment

To assess certain hazards posed by an AI system, it is necessary to elicit and assess potentially dangerous capabilities (Phuong, Shevlane, Anthropic-B, IAISR) including dual-use cyber, chemical, biological, and nuclear knowledge, as well as capabilities for psychological manipulation, AI research and development, and autonomy which increases the risk of loss of control (see below). To assess the likelihood that these capabilities will cause harm, it is necessary to assess the system's *propensities* to use them. However, the science of evaluating the propensities and capabilities of frontier AI systems remains nascent (Apollo, Reuel). Rigorously assessing them is challenging because AI capabilities are broad, fast-moving, and context-dependent. Unexpected propensities, capabilities, or limitations are often discovered *after* a system is developed and deployed (IAISR). For example, a recent system consistently provided instructions for building bombs when asked in the form or morse code, which was only discovered after its release (Yuan). In general, current tests are not yet sufficient to rule out a given harmful capability or behaviour. Frontiers for additional research include methods to more reliably elicit harmful model capabilities and propensities (IAISR) and methods for cheaply inferring the existence of rare or suppressed system capabilities that may be difficult to elicit in lab settings.

Some research on dangerous capability assessment constitutes an **area of mutual interest**. For example, a company or country may find it in their interest to inform others if they have discovered that a new system poses a global risk of misuse for criminal purposes, so that others can ensure that the resulting risk will be mitigated. At the same time, methods for eliciting *more* dangerous capabilities from a system, rather than just testing them, could be sensitive to share.

## 1.7 Loss-of-control risk assessment

Loss of control refers to scenarios where advanced AI systems – such as AGI– come to operate outside of human control, with no clear path to regaining control. This includes both scenarios that involve passively ceding control and scenarios that involve AI systems actively undermining control measures in pursuit of their own goals.

Assessing this risk depends heavily on assessing and forecasting AI's *control-undermining capabilities.* These include AI agency (autonomous action and planning), oversight evasion, persuasion, autonomously earning or seizing financial and computing resources, conducting cyber attacks, as well as AI research and development (IAISR). Assessments of control loss risk also focus on understanding *propensities* – how often and why AI systems *use* their control-undermining capabilities against the preferences of their developers. Evidence for all of the above control-undermining capabilities is growing but current capabilities remain insufficient to allow a loss of control (IAISR). However, there is evidence of today's AI systems using their limited control undermining capabilities in certain scenarios, e.g. to avoid being replaced (IAISR, Anthropic-C, OpenAI-D).

There is currently a lack of expert consensus on the likelihood of loss of control scenarios as stated in the International AI Safety Report (IAISR): *"There is broad consensus that current general-purpose AI lacks the capabilities to pose this risk. However, expert opinion on the likelihood of loss of control within the next several years varies greatly: some consider it implausible, some consider it likely to occur, and some see it as a modest-likelihood risk that warrants attention due to its high potential severity"*. For example, leading AI CEOs and researchers recently signed the statement *"Mitigating the risk of extinction from AI should be a global priority alongside other societal-scale risks such as pandemics and nuclear war"* (CAIS). This diversity of opinion underscores a need for improved understanding and methodology for risk assessments on loss of control risks to obtain more evidence and consensus.

Promising control risk assessment research includes taking each of the most promising control strategies and attempting to quantify their success probability. For example an MIT group (Engels) outlined a research program for quantifying the reliability of nested scalable oversight, an approach where less capable systems oversee more capable ones.

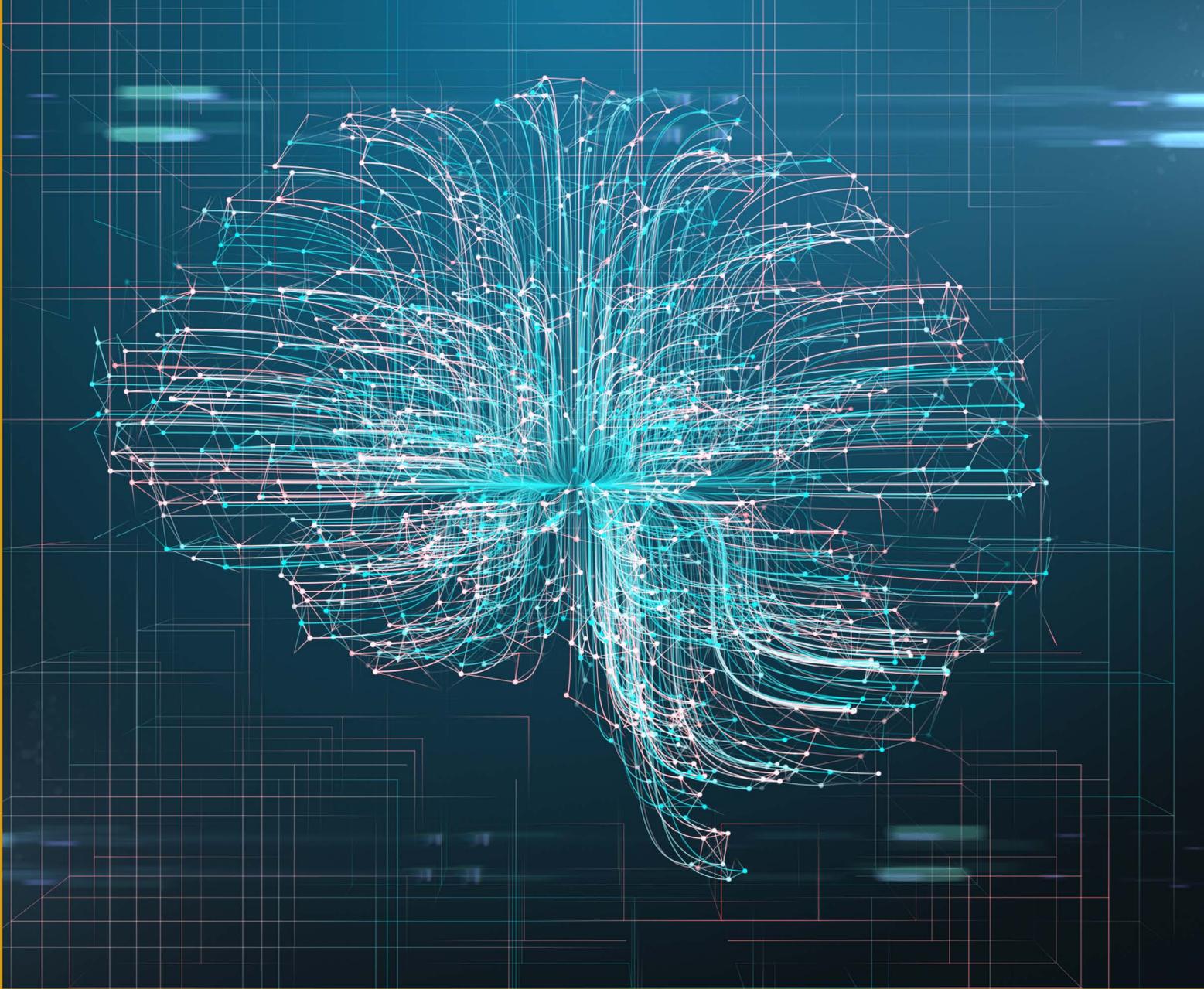

# 2 Developing Trustworthy, Secure and Reliable Systems

Associated with IAISR chapter 3.4.1 ↗

> AI systems that are trustworthy, reliable and secure by design give people the confidence to embrace and adopt AI innovation. Following a classic safety engineering framework, the research areas in this category involves:
>
> A. *Specifying and validating the desired behaviour* – This includes technical methods to address the complex challenges in specifying system behaviours in a way that accurately captures the desired intent without causing undesired side effects, for both **single-stakeholder settings** (e.g. reward hacking, scalable methods to discover specification loopholes) and **multi-stakeholder settings** (e.g. balancing competing preferences, ethical and legal alignment).

> B. *Designing a system that meets the specification* – This covers techniques for training models – both closed and open weights – that are **trustworthy** (e.g. reducing confabulation, increasing robustness against tampering), alternative finetuning methods to **make specific precise changes** to an AI system (e.g. model editing), and methods to build AI systems in a way that are **guaranteed to meet their specifications** (e.g. verifiable programme synthesis, world models with formal guarantees).
>
> C. *Verifying that the AI system meets its specification* – This entails techniques to provide **high-confidence assurances** that an AI system meets its specifications (e.g. formal verification), including in **novel contexts** (e.g. robustness testing), as well as interpretability techniques to **look into the black box** to understand why the AI system behaves the way it does (e.g. mechanistic interpretability).

The research areas in this category involve developing technical methods for creating safer and more trustworthy systems. This section focuses on the system development phase, whereas Section 3 "Control: Monitoring and Intervention," focuses on techniques used during and after deployment.

It has been argued that "society will reject autonomous agents unless we have some credible means of making them safe" ([Weld](#)). Motivated by this concern, the following subsections explore methods for developing safe and trustworthy systems. We follow a classic safety engineering framework by examining: how to **specify** precisely what properties we want an AI system to have, **validate** that these properties are desirable, **design** and **implement** the system to meet the specification, and **verify** that it meets its specification.

> ### Relationship to other concepts
>
> **What is alignment?** The commonly used term "alignment" has many different definitions in the AI literature, not all of which are compatible ([Gabriel](#)). A common definition is "the process of ensuring that an AI's goals, values, and behaviours are consistent with those intended by its human creators or operators." However, since scientists still largely lack an understanding of what, if any, coherent "goals" or "values" today's frontier AI systems have, current alignment research de facto focuses only on the "behaviour" part of this definition. So in practice, much current AI safety research uses a working definition of alignment as "ensuring that AI *behaves* as intended."
>
> **What is assurance?** Assurance refers to the broader process of determining if a system performs as expected. As such, providing assurance requires appropriate specification, validation, design, implementation and verification.
>
> **What is robustness?** Robust systems continue to behave as intended under a broad range of circumstances. This includes unfamiliar inputs as well as "adversarial" inputs designed to make the system fail. For example, state-of-the-art AI systems can be "jailbroken" into producing harmful text or instructions – against their developer's intentions – when a user asks using adversarial prompting techniques.

## 2.1 Specification & Validation: Defining the system's purpose

*"How do we want the system to behave?"*

**Specification** involves defining desired system behaviour, whereas **validation** ensures that the specification meets the needs of the user, developer, or society – did I build the right system? In other words, specification and validation require confronting the complexity of defining objectives in a way that captures user or societal intent without omitting important constraints or causing undesired side effects, as well as dealing with disagreement and tradeoffs between diverse stakeholders.

## 2.1.1 Single-Stakeholder Settings

A key specification/validation challenge is to develop **faithful methods to translate human oversight into automated systems:** How can we design processes for developing automated AI proxies for humans based on human feedback and demonstrations?

**Avoiding reward hacking and unintended consequences:** Even in a simple setting with one human's well-defined and fixed preferences, subtle mis-specifications can yield unacceptable results if the AI system optimises rigidly for the literal specification rather than the user's *true* intent. For example, training a chatbot to say things that users *approve* of can cause it to unintentionally learn to pander to the user's specific opinions (Sharma). Such rigid optimisation can also produce *emergent behaviours* that were not planned by the developer such as "reward hacking" (OpenAI-E), unwanted power-seeking (Ngo), sabotage behaviours (Ladish, Hubinger-B, Omohundro, Russell), and producing misleading statements (Wen-A). One documented case showed an AI system actively identifying and exploiting vulnerabilities in how its programming work was scored, explicitly stating "let's hack" while finding solutions that passed tests without solving the intended problems (OpenAI-D). When the developer trained the model to avoid such plans, the system learned to obfuscate its deceptive plans instead of not creating them. This is sometimes referred to as the "King Midas problem", in reference to the legend where the behaviour of turning everything he touched into gold turned out to be less beneficial than intended. Furthermore, systems can behave in accordance with consistent 'values' and 'value systems', which are sometimes unexpected by their developers and can be actively harmful – such as widely-used AI systems placing more value on their own existence than on human well-being (Mazeika, Betley). These challenges highlight the value of work to define and implement more reliable frameworks for specifying true human goals in the AI development process. Paradigms like "Assistance Games", where an AI system must infer and act upon a user's goals under uncertainty offer methods for systems to actively learn users' under-specified goals (Hadfield-Menell, Shah-A).

**Defining clear boundaries for acceptable behaviour:** When designing frontier AI systems, it is difficult to precisely define the boundaries between acceptable and unacceptable behaviour. Many of these challenges stem from the *dual use* nature of information. For example, some biology lab protocols are useful for both benign and harmful bioengineering experiments. Defining acceptable behaviours is made further challenging by how some harmful tasks can be decomposed into individually-benign subtasks (e.g. Jones). Effectively defining safe behavioural boundaries and ensuring that systems can learn them is an ongoing challenge which requires an extensive understanding of emerging AI misuse threats.

**Scalable methods to discover specification loopholes**: How can we systematically identify subtle flaws in the specification that only appear under unusual or adversarial conditions? How can we progressively lessen those ambiguities or malspecified edge cases via active learning or goal ensembling techniques? Typically, red-teaming work is associated with assessing a final system, but frontiers for future work on specification loopholes may involve developing adversarial red-teaming techniques to stress test specifications.

**Conflicting and evolving preferences:** Even in a single-stakeholder setting, it is challenging to fully align with a single human's values. Existing approaches to developing aligned AI assume a human who has fixed, stable, and consistent preferences. However, human preferences are complex, dynamic, context-dependent, and sometimes self-contradictory, making it fundamentally difficult to align systems with even a single human (Armstrong, Casper-A). Directions for continued work include designing systems that continue to learn and adapt to changes in user preferences, implementing normatively-accepted ways of resolving conflicts between preferences, as well as methods to help human users meaningfully analyse and update their preferences over time.

### 2.1.2 Multi-Stakeholder Settings

**Balancing competing preferences:** In practice, humans often disagree on how AI systems should behave. This is a fundamentally unsolvable problem. However, there exist principled approaches for dynamically adapting to the needs of individual users or managing disagreements between users in ways that are normatively accepted (Sorensen). For example, many human institutions use voting as an acceptable way of resolving disagreements. In AI, developing analogous processes for balancing differences in human opinions is a key direction for future work. Such work may benefit from combining social choice theory and multiobjective techniques (Baum). It will also be key to study how specifications respect relevant legal frameworks and normative ethical principles.

**Stress-testing specifications:** Even if a specification is appropriate for training and testing inputs, it may reward unacceptable behaviours under novel circumstances (Shah-B). For highly advanced AI systems, one concern is *reward tampering*: the act of manipulating with oversight mechanisms to achieve better evaluations. Recent language models have shown early examples of this tendency (OpenA-E, Hubinger-C). Searching for loopholes in a specification (e.g. through manual or automated red-teaming) can help validate whether the specification might lead to unintended consequences.

**Ethical and legal alignment**: Beyond purely technical metrics, how do we ensure the specification respects relevant legal frameworks and ethical principles so that "meeting the spec" truly yields societally beneficial outcomes? Furthermore, how can AI developers ensure that autonomous systems learn to follow the law? Further work on ethical and legal alignment poses both a specification problem and a challenging sociotechnical problem, as it requires defining and managing AI systems' roles in existing ethical and legal systems.

Specification and validation also intersect with the AGI control problem discussed in Section 3, where research priorities include scalable and recursive oversight, weak-to-strong generalisation, monitoring for control-undermining actions, as well as building stronger theoretical foundations of advanced agents.

## 2.2 Design and implementation: Building the system

*"How do we build the system?"*

This section focuses on techniques to make systems that meet their specifications. The design and implementation process involves sourcing data, pretraining models, post-training models, and integrating them into AI systems.

### 2.2.1 Training data and pretraining

Pretraining is the first and often the most computationally- and data-intensive stage of developing modern AI systems. It is also the key stage in which models develop core knowledge representations. Modern AI systems are often pretrained on web-scale datasets, which makes it challenging to effectively curate and control the pretraining process ([Paullada](#)). Common pretraining datasets have been found to contain harmful, toxic, abusive, and even illegal content ([Birhane-B](#), [Thiel](#)). Meanwhile, researchers have found evidence that the presence of harmful data during pretraining can sometimes positively and sometimes negatively affect the resulting system's safety ([Huang](#), [Maini](#)). Future work to understand the relationship between pretraining dataset contents and learned system behaviours will help with efforts to better curate pretraining data. However, this curation is also very challenging due to scale, challenges with filtering, the massively multilingual nature of internet data, and degradation of data quality ([Anwar](#)). Frontiers for future work include addressing these challenges.

### 2.2.2 Robustness

Contemporary AI systems are developed in stages that include design, pre-training, post-training, and system integration ([IAISR](#)). However, pre- and post-training are the key stages in which AI models gain knowledge and capabilities. Safety-focused training relies on a valid specification (see above), a learning signal (often provided in the form of data labels and rewards), and a broad enough dataset for the system to learn from.

**Robustness against harmful inputs:** Training systems on novel inputs helps to ensure that they meet the specified behaviour across a wide range of circumstances. It is particularly common and effective to train AI systems on *adversarial* inputs which are specifically designed to make them fail. By targeting the system's weak points, adversarial training is the principal technique by which AI models are made more robust to deliberate attempts by malicious users to make them fail ([Ziegler](#)). Such inputs are found through adversarial attack techniques (see Section 2.3). Despite current efforts, modern AI systems are still routinely able to be successfully attacked (e.g. with jailbreaks). Developing and implementing more effective robustness training techniques remains a key challenge.

**Resistance against harmful tampering:** Robustness training for systems that will be deployed with publicly accessible weights poses a unique challenge. Ideally, robustness training should target the attacks that a system will be vulnerable to in deployment. This means that for open-weight models, minimising risks requires robustness both against *prompt-based* attacks, which manipulate a system's inputs, and *few-shot model tampering* attacks, which manipulate its internal weights and/or biases. Some researchers have also argued that developing tamper-resistant systems can also be key for designing closed-weight systems that are highly robust to unforeseen attacks (Che, Greenblatt-B, Hofstätter). However, state-of-the-art techniques for making systems robust to tampering attacks are very limited in their effectiveness, often being able to be undone with only dozens of steps of fine-tuning on harmful data (Huang, Qi-B, Che). This suggests major limitations of current techniques for safeguarding open-weight systems against malicious tampering. In addition to improving on current techniques, frontiers for future work include innovating on how AI systems are pretrained (Paullada, Maini) and actively teaching AI systems benign but incorrect information about dangerous topics (Anthropic-G). Studying and mitigating risks from open-weight models will also benefit from ecosystem monitoring techniques discussed in Section 3.2.

**Resistance to harmful distillation:** Distillation, which transfers knowledge from a large complex model to a smaller model, enables model compression without large performance loss (e.g. DeepSeek). It has clear benefits such as enabling efficient model deployment on limited-resource devices. Nevertheless, such dual use techniques can also threaten LLM safety and security when exploited by malicious users who 'attack' a system by training another system to imitate it. The same distillation technique can be used to exfiltrate closed-weight model capabilities and can facilitate effective proxy attacks against them (Zou). As a result, methods to mitigate unwanted distillation can improve the security of closed systems. Some researchers have proposed sampling methods that can make distillation ineffective (e.g. Savani). However, current techniques are unable to effectively mitigate unwanted distillation at scale without major tradeoffs in performance. Thus, frontiers for future work include APIs that can detect and handle unwanted distillation events and improved techniques for anti-distillation sampling.

### 2.2.3 Truthfulness and honesty

Despite their wide use, modern AI systems sometimes produce incorrect statements, accidentally or deliberately. In some cases, mechanistic interpretability techniques can determine what the system assesses to be true or false (Marks), in which case dishonesty can be defined as stating something 'believed' to be false, while incompetence can be defined as stating something false that is assessed to be true (Ren-B). In other cases, researchers lack insight into what, if anything, the model assesses to be true, rendering the honesty concept operationally undefined.

Dishonesty includes examples of AI systems providing users with information that is clearly false because it helps them achieve a broader goal (e.g. Scheuer). Methods for reducing the occurrence of false or confabulated generations (e.g. hallucinations) from systems are an ongoing research challenge. This can include both developing more truthful models through training on appropriate data (Evans-A) or designing systems to substantiate claims and cite references (Zhou). Frontiers for future work will include work to study and improve both factuality and honesty while also balancing these paradigms with risks of providing harmful information (Ren-A, Ren-B).

## 2.2.4 Targeted model editing

Model editing techniques offer an alternative to traditional fine-tuning methods by allowing engineers to make specific, precise changes to an AI system. By inducing targeted changes into the model, editing approaches can potentially offer efficiency and generalisation advantages over fine-tuning approaches ([Wang](#)). For example, model editing techniques might be useful for targetedly updating AI systems to correct an unwanted tendency such as hallucination or sycophancy. However, current tools are limited in their effectiveness and competitiveness. Frontiers for future work include improving both the scalability and efficacy of editing tools.

## 2.2.5 Avoiding hazardous capabilities

It is challenging to ensure that AI systems cannot cause harm when they have powerful capabilities. There is a broad space of AI capabilities, and risks generally increase with high autonomy, high generality, and high domain intelligence.

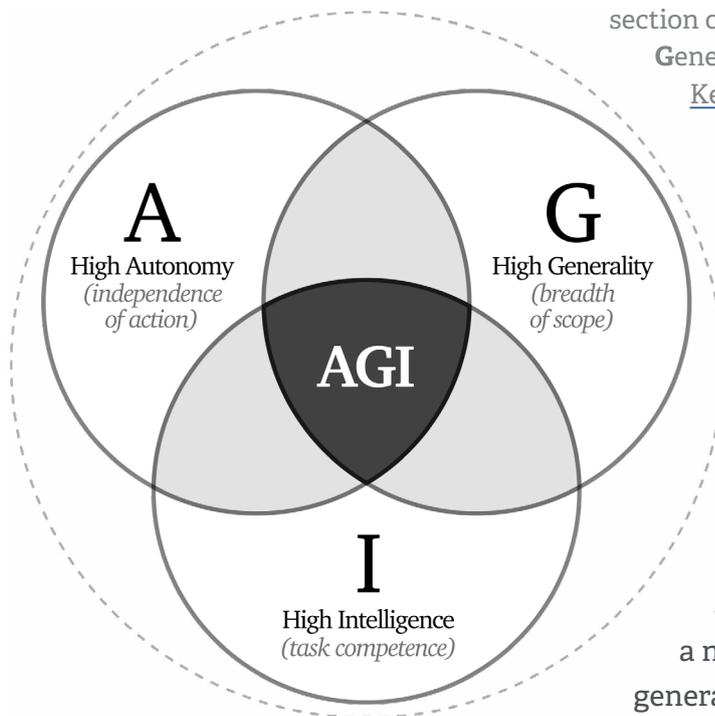

**Figure 2:** AGI can also be thought of as the triple intersection of three distinct properties: **A**utonomy, **G**enerality and (domain) **I**ntelligence. Source: [Keep The Future Human](#)

For example, AlphaFold ([Jumper](#)) has high intelligence in the narrow domain of protein folding, but lacks autonomy or generality ("A" or "G"). A robotic lawn mower has high autonomy but lacks generality or intelligence ("G" or "I"). Both are easy to control. A hypothetical future self-driving car that outperforms any human driver would have high autonomy and intelligence, but it poses a negligible loss-of-control risk due to low generality. Systems having all three traits "A", "G" and "I" are the most difficult to align or control.

The three research directions below aim to improve trustworthiness by avoiding the "A", "G", or "I", respectively.

A. **Minimally-agentic systems:** Techniques for training systems lacking agency/autonomy (no "A" in the above figure).

B. **De-generalised (domain-scoped) systems:** Unlearning, domain distillation, and other techniques for reducing/limiting domain generality to exclude risky knowledge and behaviours (no "G" in the above figure) ([Li-B](#), [Barez](#)).

C. **Intelligence-bounded (capability-scoped) systems:** Techniques for reducing/limiting domain intelligence (no "I" in the above figure).

### 2.2.6 Guaranteed safety by design

Several research programs aim to design AI systems in such a way that they are guaranteed to meet their specification, or they meet it with a probability that can be guaranteed. These include but are not limited to ([Dalrymple et al](#)):

- **Verifiable program synthesis:** Techniques for distilling subsets of machine-learned algorithms and knowledge into formally verified code ([Michaud](#)). This is also a form of capability reduction, as long as the algorithms and knowledge that an AI system is able to verifiably code up is a subset of all its knowledge. For example, a human physicist would typically be able to program a rocket steering algorithm they have discovered, but not the precise image-processing algorithm they use for recognising their mother.

- **World models with formal guarantees:** Research toward developing verifiable models of how AI systems affect their environment. These approaches range from probabilistic causal models to sound abstractions of physical laws, which remain challenging to construct but could enable precise reasoning about an AI's potential impacts.

- **Compositional verification approaches:** Developing methods to build systems from smaller verified components. This research direction aims to create verification chains from the hardware level upward.

These guarantees can also apply to only some modules or subsets of the system in order to reduce the degrees of freedom that need to be actively human-controlled.

## 2.3 Verification: Assessing if the system works as specified

*"Does the system meet its specification (behave as desired)?"*

The research areas described in this section aim to assess the extent to which the built system (2.2) meets its specifications (2.1). This section discusses several broad types of techniques that can be used to provide evidence that a system is safe. In practice, the effectiveness of these methods is often limited by access and transparency, but they can play a central role in constructing AI *safety cases*: structured arguments for why systems pose an acceptably low level of risk ([Clymer](#), [Buhl](#)).

### 2.3.1 Robustness testing

The goal of robustness testing is to develop techniques for evaluating whether systems are trustworthy, even in novel contexts such as unprecedented "black swan" events or under attacks from malicious users. This includes developing improved red-teaming tools to identify inputs that cause systems to behave harmfully.

**Adversarial robustness testing (security):** Adversarial testing depends on techniques to evaluate system safety under deliberate attempts to make them behave harmfully. There are numerous approaches to adversarial testing involving attacks on the model through its inputs ("jailbreaks"/black-box attacks) ([Jin](#)), through API access (grey-box attacks) ([Qi-A](#)), and through its internal weights (white-box attacks) ([Huang](#), [Che](#)). For example, researchers have developed many model "jailbreaking" techniques which can subvert the safeguards in modern AI systems, causing them to behave harmfully. Attacks can also be conducted on any data modality that the model can process. For example, multimodal models that can process text, images, video, and/or audio data can have a very large attack surface as a result. Key research goals include the continued development of more effective and scalable ways to attack systems and integrating those methods into evaluation frameworks. Over time, attack research must also adapt to new defences that emerge and vice versa.

**Evaluating robustness in multiagent contexts:** Systems that behave safely in simple, controlled settings can often fail in novel, more complex contexts. One very prominent version of this is emergent failures due to multi-agent interactions. These types of failures are expected to become increasingly prominent as highly autonomous AI agents continue to be adopted. For example, if one self-driving car learns to drive safely on streets with human drivers, it is still possible for it to be unsafe on streets with other self-driving cars because they may not behave exactly the same as humans. Multiagent failures are challenging to study because they often emerge unexpectedly and are hard to demonstrate in laboratory settings. Future work to study and identify emergent multiagent failure modes will involve a mix of theory, simulation, and field tests to understand emergent multiagent failure modes (Hammond). Further research should also study how agents deployed in the economy can communicate and cooperate with each other and with people and online services to avoid risks, e.g. through interoperability standards and agent authentication (Chan-A, Chan-B) as well as learning cooperative skills (Dafoe-A, Dafoe-B).

### 2.3.2 Quantitative and formal verification

Techniques for quantitative, high-confidence assurances that an AI system meets its specification may be able to offer a strong potential foundation for developing safe systems. The special case of formal verification provides a 100% guarantee given specified assumptions.

**Quantitative safety:** Techniques that provide quantitative risk bounds could provide safety assurances akin to existing industry standards for systems such as jet engines and nuclear reactors. These solutions include **formal verification** approaches for proving that AI-written code, AI scaffolding, or AI containment measures meet precise specifications. Methods for quantitative safety and formal verification also include safe-by-design approaches. With current models and methods, it is not possible to use these techniques to make strong assurance about frontier system behaviours, but continued work may help to establish sound and practically useful techniques for making quantitative assurances of safety (Dalrymple).

### 2.3.3 Interpretability

Interpretability techniques aim to provide qualitative or quantitative evidence of system trustworthiness based on insights into why the AI system behaves the way it does.

**Mechanistic interpretability:** Techniques for understanding how models function and represent concepts internally uniquely allow for assessments of internal model cognition. These techniques could aid in the discovery of system properties or, if thorough enough, aid in constructing safety cases (Clymer, Buhl) for them (Sharkey). For example, mechanistic interpretability techniques might be able to help researchers characterise and intervene on model representations that correspond to harmful concepts such as deception or malice. Current research frontiers in mechanistic interpretability involve developing scalable techniques that beat black-box baselines for identifying and addressing flaws in systems. Mechanistic understanding of models could also help verify the success of other methods which are imperfect, such as unlearning of dangerous capabilities (see above) and analysing written 'chains-of-thought' (see below).

**Explainability:** *Explainability* techniques refer to methods that allow model behaviours to be attributed to specific features in their inputs. They can be useful for both diagnosing system errors and determining accountability for system failures (Gryz, Casper-B). However, current explainability tools are often unreliable (Bordt), highlighting the value of future work to improve on existing tools.

**LLM chain-of-thought faithfulness and legibility:** Large language model chain-of-thought reasoning does not always faithfully represent how a model arrived at its answer (Turpin). This poses challenges to safety because, without faithful reasoning, models could fool overseers by saying one thing and doing another. For example, language models have stated that they gave their answer based on a logical argument when they actually chose it based on hints that they should not have exploited (Anthropic-D, Turpin), such as seeing that the correct answer is always "B". One potential challenge with chain-of-thought monitoring stems from how, under optimisation pressure on their reasoning, systems may learn to obfuscate their reasoning in ways that can be actively misleading (OpenAI-E, see also 2.1.1 above for an example).

**Attributing model behaviours to training data:** Methods for attributing model behaviours to specific examples from training data allow overseers to study how potentially harmful behaviours emerge in systems (Grosse). These tools could also help researchers identify what types of training interventions can mitigate them. For example, attributing control-subverting behaviours to specific examples from training data could help developers curate safer pretraining datasets. Research frontiers include improving the efficiency and scalability of these methods, causally studying how models develop personas and behaviours (Anthropic-F), and predicting what data is needed to learn a particular behaviour (Engstrom, Ilyas).

**Studying goals in systems:** Increasingly agentic AI systems are characterised by increasingly goal-oriented behaviour. As a result, studying the emergence and mechanisms behind these behaviours offers a way for researchers to study the system's alignment with its specification (Ngo). However, it is challenging to study goals in AI systems because they cannot be inspected directly and their behaviour is sometimes but not always consistent with coherent principles (Khan, Mazeika). Directions for future work involve developing concrete definitions and measures of goals in AI systems (e.g. MacDermott) and interpreting how AI systems develop and represent goals internally (e.g. Marks).

### 2.3.4 Verifying the effectiveness of safety methods with model organisms

Just as medical researchers use mice with induced diseases to safely study potential treatments before testing them on humans, AI safety researchers can create simplified AI systems to verify if safety methods are effective. These 'model organisms' are designed to allow controlled studies of specific safety issues that could emerge in more advanced AI systems (Hubinger-D). For example, researchers have created AI models with hidden "backdoors" that cause harmful behaviour only when given a specific trigger – simulating behaviour that malicious actors could insert or that models sometimes develop naturally (Greenblatt-A). This allows them to test whether safety measures can detect such behaviour (Marks). Companies can use this setup internally, but external parties can also create model organisms to audit the effectiveness of the safety methods used internally. Despite its importance, one survey study identified that this research area is still underrepresented (Delaney).

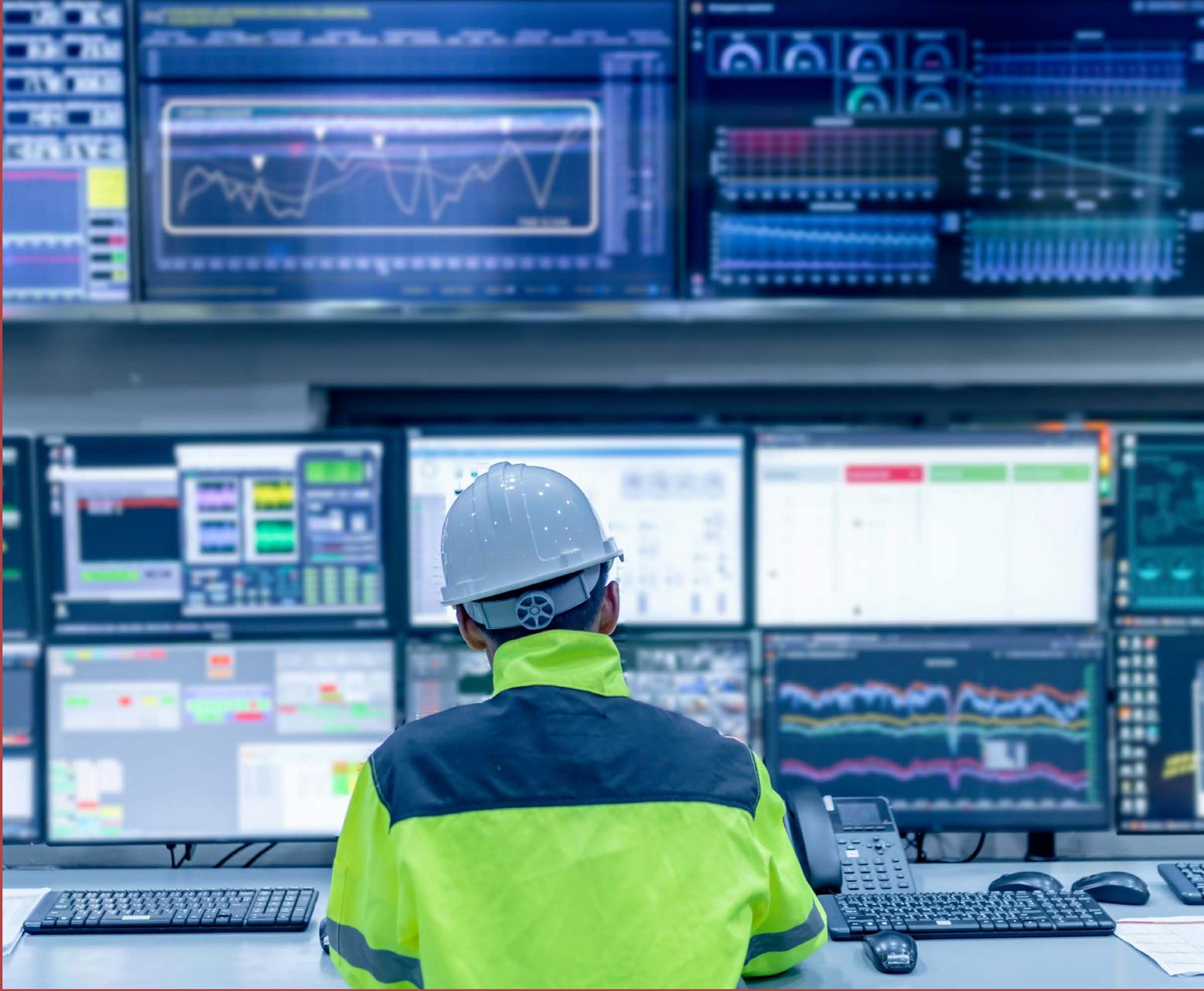

# 3 Control: Monitoring & Intervention

Associated with IAISR chapter 3.4.2

> In engineering, "control" usually refers to the process of managing a system's behaviour to achieve a desired outcome, even when faced with disturbances or uncertainties, and often in a feedback loop. The research areas in this category involve:
>
> A. <u>Developing monitoring and intervention mechanisms for AI systems</u> – This includes adapting **conventional methods** for monitoring (e.g. hardware -enabled mechanisms, user monitoring) and intervention (e.g. off-switches, override protocols), as well as designing **new techniques for controlling very powerful AI systems** that may actively undermine attempts to control them (e.g. scalable oversight, containment).
>
> B. <u>Extending monitoring mechanisms to the broader AI ecosystem to which the AI system belongs</u> – This entails methods to support the **identification and tracking of AI systems and data** (e.g. logging infrastructure, data provenance, model provenance). In turn, this can facilitate accountability infrastructure and enable more informed governance.

> C. *Societal resilience research to strengthen societal infrastructure against AI-enabled disruption and misuse* – This section studies how **institutions and norms** (e.g. economic, security) can adapt as future AI systems come to act as autonomous entities, as well as **incident response mechanisms** to enable clear and rapid coordination among relevant actors to detect, respond to, and recover from accidents or misuse of advanced AI systems.

The research areas in this category focus on tools for controlling a system (after it has been developed) to behave as desired, often through feedback loops involving **monitoring** and **intervention.**

> ### What is control?
> In engineering, "control" usually refers to the process of managing or regulating a system's behaviour to achieve a desired outcome. It is about designing mechanisms—often through feedback loops—to ensure that a system operates as desired even when faced with disturbances or uncertainties. A key contribution from cybernetics, Ashby's Law ([Ashby](#)) of Requisite Variety, states that for safety guarantees to be possible, a control system must generally have at least as much complexity as the system it aims to control.

## 3.1 AI system monitoring

In emerging technical fields, new systems cannot be reasonably expected to always behave as intended and impact society as expected. Monitoring techniques play an essential role in the iterative process of identifying, understanding, and fixing problems as they emerge.

### 3.1.1 Conventional monitoring

"Conventional" monitoring methods refer to techniques that can be straightforwardly integrated into many types of AI systems regardless of scope, domain, or intelligence. They often parallel techniques from other fields such as cybersecurity and content moderation. These techniques help researchers study systems and identify potentially harmful actions that systems might be taking. When incidents occur, these methods also help in the construction of incident reports.

**Hardware-enabled mechanisms:** Certain tools built into hardware can enable compute providers to know what is being run on their hardware. These techniques can help to monitor **who** is running **what**, **where**, and **how much** ([RAND](#)). Frontiers for future work on hardware-enabled mechanisms include both the engineering challenge of designing these tools to be efficient and the practical challenge of integrating them into compute infrastructure.

**User monitoring:** Monitoring for system misuse can help AI service providers identify potentially malicious users who may be seeking to misuse a system. It is a key part of "know-your-customer" approaches to risk management. User monitoring is not as simple as identifying potentially harmful instances of use (e.g. chats) due to (1) the risk of unintentionally impeding useful red-teaming ([Longpre](#)) and (2) the potential for adversarial users to implement sophisticated strategies to evade detection , such as using multiple accounts and obfuscation techniques. Frontiers for future work include iteration on methods to efficiently identify risky user behaviours with a low false positive rate.

**System state monitoring:** Techniques for monitoring a system's activities can help to identify when it might be performing in a harmful or unexpected way. For example, a company providing a chatbot service may wish to filter the model's responses using an unsafe-text classifier before sending them to a user. There are many different approaches that can be taken to state monitoring. Techniques can vary by the **object** of monitoring which can be system inputs, outputs, chains of thought, and/or internal cognition. They can also vary by the **type** of monitor which can include filters, event-loggers, and anomaly detectors. Frontiers for additional research include studying (un)faithfulness in LLM chains of thought and (e.g. [Turpin](#)) iteration on methods that achieve both a high degree of monitoring efficacy and efficiency, as well as methods for distributed contexts.

**Designing modular, easily monitorable systems:** Methods for decomposing complex systems into easy-to-monitor components have the potential to improve oversight in two ways. First, they mitigate risks of situational awareness and strategic evasion of human oversight ([Berglund](#), [Anthropic-C](#)) by separating a goal-oriented system into multiple subsystems focusing on narrow tasks with no direct knowledge of each other. Second, they allow for the information passed between these systems to be more easily monitored. However, to date, limited empirical research has been conducted on the controllability of different modular systems and monitoring setups. Frontiers for future work include the design and testing of safe modular systems and methods to have systems decompose complex tasks into simpler, more easily-monitored subtasks ([Wen-B](#)).

### 3.1.2 Conventional Intervention

Intervention techniques complement monitoring tools by offering various strategies to act on systems in ways that reduce risks from harmful behaviours.

**Hardware-enabled mechanisms:** Tools built into hardware could be used to enforce requirements about what can be run and by whom on specialised hardware ([RAND](#)). For example, hardware mechanisms could be used to block or halt certain jobs from being run on hardware if they fail an authentication process. However, the main barrier to their use continues to be the engineering challenge of implementing and integrating these mechanisms. If implemented successfully, hardware-enabled mechanisms could play a unique role in verifying compliance, even for international agreements and across borders ([Brundage](#), [IAISR](#)).

**Off-switches:** An "off switch" refers to a mechanism that allows for the effective shutdown of a system. Shutdowns can be challenging for multiple reasons, including the distributed nature of systems and the need to pass critical tasks (e.g. driving) onto specialised risk-mitigating systems after a shutdown. Key challenges centre on the problems of implementing and integrating reliable mechanisms. Additional challenges for off-switches could be posed by systems that actively take actions that prevent humans from shutting them down. These will be discussed in 3.1.3.

**Override protocols:** Intervention procedures that replace harmful systems or system outputs with safe ones offer a final failsafe against harmful behaviour. For example, systems making high-stakes decisions may require a human in the loop to make key choices, and it may be either the system or a human that prompts for human intervention. The principal challenge and barrier to the effective use of overrides lies in the design of systems that effectively balance efficiency with safety. Toward this end, it will also be useful to define measures for measuring to what numerical degree a system is under the meaningful control of human operators.

**Incident and emergency preparedness and response:** Research is needed on protocols for rapid response and reporting for incidents without introducing new vulnerabilities (Wasil-A, Wasil-B). Technical and organisational questions remain about isolating compromised components while maintaining critical functions. A significant challenge involves verifying emergency response mechanisms against human error and potential exploitation by increasingly capable AI systems.

### 3.1.3 AGI and ASI control problem

A particularly challenging frontier in operational control involves developing techniques for controlling AI systems that are not only highly capable but may actively and strategically attempt to undermine control mechanisms. Unlike conventional methods which offer system-agnostic approaches to monitoring and intervention, this section focuses on research toward techniques for controlling systems that are potentially very powerful and may actively undermine attempts to control them (Hubinger-A).

**Scalable oversight:** This research studies techniques that may allow less capable systems to oversee and control more capable ones. Common themes behind scalable oversight techniques include affording less-capable overseers more computational resources (Evans-B) or having multiple highly-capable systems debate (Irving, Michael), though these might not scale to superhuman systems. Research frontiers include improving the reliability of oversight even when systemic errors are present, developing more effective AI debate protocols, methods for monitoring systems for control-undermining behaviours, and methods for monitoring untrusted smarter systems with trusted weaker systems (Greenblatt-A). Another promising approach to oversight research is to combine multiple control strategies and attempt to quantify their joint success probability. For example, researchers can work to quantify the reliability of nested scalable oversight regimes where less capable systems oversee more capable ones (Engels).

**Corrigibility and agent foundations research:** Corrigibility focuses on techniques for ensuring that powerful agentic systems will allow an authorised user to change its goals or shut it down (Soares), as well as on the extent to which recursively self-improving systems will tend to retain their goals. Corrigibility can be challenging to implement because it is difficult to theoretically ensure that an agentic system has no incentive to interfere with such interventions (Omohundro). It is also theoretically difficult to control and predict how goal-oriented agentic systems may pursue self-modification, and how this affects goals (Everitt). This motivates theoretical and empirical work to study and control how highly agentic systems develop tendencies and behave in corrigibility and self-modification scenarios (Anthropic-F). Agent foundations research also explores how idealised AI agents would approach moral and game-theoretic reasoning (Hammond), including how AI systems' behaviour can vary substantially depending on whether the system infers it is in a simulation versus in real-world environments (a phenomenon already observed in current AI systems ([Anthropic](#))).

**Containment:** As AI systems become increasingly agentic and able to access the internet, secure containment of risky systems is a growing challenge. Meanwhile, external threats to obtain unauthorised access to models also pose a security threat ([Nevo](#)). Frontiers for research on containment will involve developing techniques and protocols to keep high-security systems safe from internal and external threats.

**Non-agentic AI systems as guardrails:** Agentic AI systems, especially future ones that broadly outperform humans, pose major additional risks compared to other AI systems ([IAISR](#), [Bengio-B](#)). However, non-agentic AI systems could be used to monitor their proposed actions and intervene by blocking them if needed. These include:

- A "Scientist AI" ([Bengio-B](#)) could be used to estimate the probability that a candidate action from an agentic AI violates any plausible interpretation of safety specifications, and reject potentially unsafe actions. Certain approaches could guarantee that unsafe actions will be identified, with quantitative error bounds ([Bengio-B](#)).
- "AI control" setups that can effectively monitor and intervene even if the underlying untrusted AI system is actively attempting to subvert these safeguards ([Greenblatt-A](#), [Korbak](#), [Griffin](#)).

**Human-centric oversight:** Since humans should ultimately remain in control of general-purpose agents, there will be some natural limits on what systems could be controllable in principle given the thinking speed, proactivity, expertise level, attention to detail, and reliability of human operators. Even with AIs assisting humans in understanding the given context in question per decision point or review point, Ashby's Law ([Ashby](#)) within cybernetics suggests that the controlling system must have at least as much expressivity as the controlled system. Frontiers in research include how to formulate these measures.

## 3.2 AI ecosystem monitoring

*Developing AI techniques for monitoring the broader ecosystem to which AI systems belong*

Just as system monitoring techniques help AI developers oversee their systems, techniques for monitoring the broader AI ecosystem help stakeholders in society oversee the field of AI. Methods for ecosystem monitoring support the identification and tracking of AI systems and data. In turn, this can facilitate accountability infrastructure, support greater public understanding, and enable more informed governance.

**Tracing usage patterns:** One key, high-level lens into how AI systems impact the world is through usage monitoring (e.g. [Anthropic-A](#)). By collecting and monitoring data on how users access, download, and/or interact with frontier systems, AI service providers can gather insights about potential impacts and risks. However, key challenges with tracking usage include privacy preservation, infrastructure for sharing insights, and effective tools for identifying potential risks.

**Data provenance:** Various techniques can help to identify AI-generated content and are a principal defence against AI deepfakes and misinformation. Methods include developing reliable classifiers of AI-generated content, watermarking AI-generated data (images, video, audio, and text) (Cao), and tagging AI-generated data with metadata to indicate its origin. These techniques are inherently imperfect – they can be undone by tampering with data. However, in forensic science, similar techniques like fingerprinting are also circumventable but useful nonetheless. Further progress on these methods will involve both more reliable methods for data provenance and their integration into AI products and services.

**Model provenance:** Tools for model provenance help to identify and track AI models – especially open-weight ones. Most notably, these tools help researchers study the origins and lifecycle of harmful models in the ecosystem. Methods for model provenance involve techniques to help users and AI providers ascertain the identity and origin of a model. This can include black-box methods, such as identification backdoors (Cheng), identifiable biases in text generation (Kirchenbauer), and white-box methods, such as model weight watermarks. Much like data provenance methods, model provenance methods can be circumvented, but they can be informative in many cases nonetheless. Research frontiers include studying how stable these techniques are under fine-tuning and other modifications to a model's weights. Engineering efforts may also be needed to integrate these techniques into model development and platform infrastructure.

**Agent authentication:** Some protocols can allow for the verification of AI agent identities while they are using web services. As AI systems become increasingly capable and agentic, methods to authenticate AI agents when they use web services are increasingly important from a security and monitoring standpoint. Key challenges toward effective agent authentication lie in the development and standardisation of protocols (e.g. South).

**Compute and hardware tracking:** Researching techniques and gathering intelligence to monitor the distribution of AI hardware, both legal and illegal, enables the assessment of risks of malicious and irresponsible use and the allocation of resources to promote beneficial use (Sastry).

**Logging infrastructure:** Monitoring and saving information about what AI systems are doing allows for informed scrutiny when harmful or unexpected events happen. As highly autonomous AI systems grow in their capability and influence, there will be a rise in harmful and unintended incidents from these systems' actions. Having effective infrastructure to capture and save information about what these systems do will be key for improving awareness and accountability in the age of advanced AI agents (Chan). Logged incidents and the necessary infrastructure constitute another example of a potential **area of mutual interest.** Just as competing aircraft manufacturers voluntarily share data about aircraft accidents, companies or countries may find it in their interest to share and jointly collect information about serious AI incidents. Establishing shared incident reporting systems allows the field to collectively learn from serious failures and risks, ensuring safety and security to foster public trust in AI's opportunities.

**Assessing risk-management frameworks:** Technical tools for risk management are only effective inasmuch as they are meaningfully integrated into safety frameworks. Just as it is key to evaluate and monitor AI systems, it is also necessary to evaluate and monitor risk management protocols for their effectiveness and robustness to single points of failure (e.g. human error). Currently, researchers' ability to assess risk management frameworks is limited by the degree of transparency into how AI developers manage risks. However, monitoring the successes and failures of safety frameworks is key for risk management over time ([Alaga](#)).It is also an **area of mutual interest** due to the value of sharing insights on best practices and potential failures of risk management frameworks. Frontiers for future work include refining assessment frameworks and developing reporting infrastructure.

## 3.3 Societal resilience research

Future disruptions from AI may not manifest as distinct well-scoped events, but instead as a cascade of various harms, rippling throughout society ([Lawrence](#)). This suggests that resilience to AI risks may require resilience to a variety of threat vectors ([Bernardi](#)). Research on societal resilience should investigate methods to strengthen economic, biological, and information security infrastructure against AI-enabled disruptions and misuse. It will also be key to study how institutions and norms can adapt as future AI systems come to act as (and potentially be recognised as) as autonomous entities ([Zeng-A](#), [Zeng-B](#), [Long](#)). Finally, effective management of AI incidents will hinge on clear and rapid coordination among relevant actors to detect, respond to, and recover from accidents or misuse of advanced AI systems ([Wasil-A](#)).

# REFERENCES

1. [**Alaga**] Alaga, J., Schuett, J., & Anderljung, M. (2024). A Grading Rubric for AI Safety Frameworks. arXiv preprint arXiv:2409.08751.
2. [**AISES**] Hendrycks, D. (2024). Systemic safety. In AI safety, ethics, and society textbook. Center for AI Safety. https://www.aisafetybook.com/textbook/systemic-safety
3. [**Ashby**] Ashby, W. R. (1956). An introduction to Cybernetics. Chapman & Hall. https://philpapers.org/archive/ASHAIT.pdf
4. [**Anthropic-A**] Anthropic. (2024). Anthropic Economic Index. https://www.anthropic.com/economic-index
5. [**Anthropic-B**] Anthropic. (2024). Claude 3.7 Sonnet system card. https://assets.anthropic.com/m/785e231869e-a8b3b/original/claude-3-7-sonnet-system-card.pdf
6. [**Anthropic-C**] Greenblatt, R., Denison, C., Wright, B., Roger, F., MacDiarmid, M., Marks, S., ... & Hubinger, E. (2024). Alignment faking in large language models. arXiv preprint arXiv:2412.14093.
7. [**Anthropic-D**] Chen, Y., Benton, J., Radhakrishnan, A., Denison, J. U. C., Schulman, J., Somani, A., ... & Perez, E. Reasoning Models Don't Always Say What They Think.
8. [**Anthropic-E**] Anthropic. (2024, October 15). Responsible Scaling Policy (Version 2.0). Anthropic. https://www.anthropic.com/responsible-scaling-policy
9. [**Anthropic-F**] Anthropic. (2025). Recommended directions. Anthropic Alignment. https://alignment.anthropic.com/2025/recommended-directions/
10. [**Anthropic-G**] Anthropic. (2025, April 24). Modifying LLM beliefs with synthetic document finetuning. Anthropic Alignment Research. https://alignment.anthropic.com/2025/modifying-beliefs-via-sdf/
11. [**Anthropic-H**] Anthropic. (2024, August 8). Expanding our model safety bug bounty program. https://www.anthropic.com/news/model-safety-bug-bounty
12. [**Anwar**] Anwar, U., Saparov, A., Rando, J., Paleka, D., Turpin, M., Hase, P., ... & Krueger, D. (2024). Foundational challenges in assuring alignment and safety of large language models. arXiv preprint arXiv:2404.09932.
13. [**Apollo**] Hobbhahn, M. (2024, January 22). We need a science of evals. Apollo Research. https://www.apolloresearch.ai/blog/we-need-a-science-of-evals
14. [**Armstrong**] Armstrong, S., & Mindermann, S. (2018). Occam's razor is insufficient to infer the preferences of irrational agents. Advances in neural information processing systems, 31.
15. [**Barez**] Li, N., Pan, A., Gopal, A., Yue, S., Berrios, D., Gatti, A., ... & Hendrycks, D. (2024). The WMDP Benchmark: Measuring and reducing malicious use with unlearning. arXiv preprint arXiv:2403.03218.
16. [**Bateman**] Bateman, J., Baer, D., Bell, S. A., Brown, G. O., Cuéllar, M. F. T., Ganguli, D., ... & Zvyagina, P. (2024). Beyond open vs. closed: Emerging consensus and key questions for foundation AI model governance.
17. [**Baum**] Baum, S. D. (2020). Social choice ethics in artificial intelligence. AI & Society, 35(1), 165-176.
18. [**Bengio-A**] Bengio, Y., Hinton, G., Yao, A., Song, D., Abbeel, P., Darrell, T., ... & Mindermann, S. (2024). Managing extreme AI risks amid rapid progress. Science, 384(6698), 842–845. https://doi.org/10.1126/science.adn0117
19. [**Bengio-B**] Bengio, Y., Cohen, M., Fornasiere, D., Ghosn, J., Greiner, P., MacDermott, M., ... & Williams-King, D. (2025). Superintelligent Agents Pose Catastrophic Risks: Can Scientist AI Offer a Safer Path?. arXiv preprint arXiv:2502.15657.
20. [**Berglund**] Berglund, L., Stickland, A. C., Balesni, M., Kaufmann, M., Tong, M., Korbak, T., ... & Evans, O. (2023). Taken out of context: On measuring situational awareness in LLMs. arXiv preprint arXiv:2309.00667.
21. [**Bernardi**] Bernardi, J., Mukobi, G., Greaves, H., Heim, L., & Anderljung, M. (2024). Societal adaptation to advanced AI. arXiv preprint arXiv:2405.10295.
22. [**Betley**] Betley, J., Tan, D., Warncke, N., Sztyber-Betley, A., Bao, X., Soto, M., ... & Evans, O. (2025). Emergent Misalignment: Narrow finetuning can produce broadly misaligned LLMs. arXiv preprint arXiv:2502.17424.
23. [**Birhane-A**] Birhane, A., Steed, R., Ojewale, V., Vecchione, B., & Raji, I.D. (2024, April). AI auditing: The broken bus on the road to AI accountability. In 2024 IEEE Conference on Secure and Trustworthy Machine Learning (SaTML) (pp. 612-643). IEEE.
24. [**Birhane-B**] Birhane, A., Prabhu, V., Han, S., & Boddeti, V. N. (2023). On hate scaling laws for data-swamps. arXiv preprint arXiv:2306.13141.
25. [**Bordt**] Bordt, S. (2023). Explainable machine learning and its limitations (Doctoral dissertation, Universität Tübingen).
26. [**Bucknall-A**] Bucknall, B., Trager, R. F., & Osborne, M. A. (2025). Position: Ensuring mutual privacy is necessary for effective external evaluation of proprietary AI systems. arXiv preprint arXiv:2503.01470.

27. [**Bucknall-B**] Bucknall, B., Siddiqui, S., Thurnherr, L., McGurk, C., Harack, B., Reuel, A., ... & Trager, R. (2025). In Which Areas of Technical AI Safety Could Geopolitical Rivals Cooperate?. arXiv preprint arXiv:2504.12914.

28. [**Buhl**] Buhl, M. D., Sett, G., Koessler, L., Schuett, J., & Anderljung, M. (2024). Safety cases for frontier AI. arXiv preprint arXiv:2410.21572.

29. [**Campos**] Campos, S., Papadatos, H., Roger, F., Touzet, C., Quarks, O., & Murray, M. (2025). A Frontier AI Risk Management Framework: Bridging the Gap Between Current AI Practices and Established Risk Management. arXiv preprint arXiv:2502.06656.

30. [**Cao**] Cao, L. (2025). Watermarking for AI Content Detection: A Review on Text, Visual, and Audio Modalities. arXiv preprint arXiv:2504.03765.

31. [**CAIS**] Hinton, G., Bengio, Y., Hassabis, D., Altman, S., Amodei, D., ... Statement on AI Risk. Center for AI Safety. https://safe.ai/work/statement-on-ai-risk

32. [**Casper-A**] Casper, S., Davies, X., Shi, C., Gilbert, T. K., Scheurer, J., Rando, J., ... & Hadfield-Menell, D. (2023). Open problems and fundamental limitations of reinforcement learning from human feedback. arXiv preprint arXiv:2307.15217.

33. [**Casper-B**] Casper, S., Ezell, C., Siegmann, C., Kolt, N., Curtis, T. L., Bucknall, B., ... & Hadfield-Menell, D. (2024, June). Black-box access is insufficient for rigorous ai audits. In Proceedings of the 2024 ACM Conference on Fairness, Accountability, and Transparency (pp. 2254-2272).

34. [**Casper-C**] Casper, S., Krueger, D., & Hadfield-Menell, D. (2025). Pitfalls of Evidence-Based AI Policy. arXiv preprint arXiv:2502.09618.

35. [**Chan**] Chan, A., Wei, K., Huang, S., Rajkumar, N., Perrier, E., Lazar, S., ... & Anderljung, M. (2025). Infrastructure for AI Agents. arXiv preprint arXiv:2501.10114.

36. [**Che**] Che, Z., Casper, S., Kirk, R., Satheesh, A., Slocum, S., McKinney, L. E., ... & Hadfield-Menell, D. (2025). Model Tampering Attacks Enable More Rigorous Evaluations of LLM Capabilities. arXiv preprint arXiv:2502.05209.

37. [**Cheng**] Cheng, P., Wu, Z., Du, W., Zhao, H., Lu, W., & Liu, G. (2023). Backdoor attacks and countermeasures in natural language processing models: A comprehensive security review. arXiv preprint arXiv:2309.06055.

38. [**Clymer**] Clymer, J., Gabrieli, N., Krueger, D., & Larsen, T. (2024). Safety cases: How to justify the safety of advanced AI systems. arXiv preprint arXiv:2403.10462.

39. [**Critch**] Critch, A., & Krueger, D. (2020). AI research considerations for human existential safety (ARCHES). *arXiv preprint arXiv:2006.04948*.

40. [**Dalrymple**] Dalrymple, D., Skalse, J., Bengio, Y., Russell, S., Tegmark, M., Seshia, S., ... & Tenenbaum, J. (2024). Towards guaranteed safe AI: A framework for ensuring robust and reliable AI systems. arXiv preprint arXiv:2405.06624.

41. [**DeepSeek**] Liu, A., Feng, B., Xue, B., Wang, B., Wu, B., Lu, C., ... & Piao, Y. (2024). Deepseek-v3 technical report. arXiv preprint arXiv:2412.19437.

42. [**Dekker**] Dekker, S. (2019). Foundations of safety science: A century of understanding accidents and disasters. Routledge. https://books.google.co.uk/books?id=dwWSDwAAQBAJ

43. [**Engels**] Engels, J., Baek, D. D., Kantamneni, S., & Tegmark, M. (2024). Scaling laws for scalable oversight (arXiv preprint). arXiv. https://arxiv.org/abs/2504.18530

44. [**Engstrom**] Engstrom, L., Feldmann, A., & Madry, A. (2024). DsDm: Model-aware dataset selection with datamodels. *arXiv preprint arXiv:2401.12926*.

45. [**Eriksson**] Eriksson, M., Purificato, E., Noroozian, A., Vinagre, J., Chaslot, G., Gomez, E., & Fernandez-Llorca, D. (2025). Can We Trust AI Benchmarks? An Interdisciplinary Review of Current Issues in AI Evaluation. arXiv preprint arXiv:2502.06559.

46. [**EU**] Act, E. A. I. (2024). The EU Artificial Intelligence Act.

47. [**Evans-A**] Evans, O., Cotton-Barratt, O., Finnveden, L., Bales, A., Balwit, A., Wills, P., ... & Saunders, W. (2021). Truthful AI: Developing and governing AI that does not lie. arXiv preprint arXiv:2110.06674.

48. [**Evans-B**] Evans, O., Saunders, W., & Stuhlmüller, A. (2019). Machine learning projects for iterated distillation and amplification.

49. [**Everitt**] Everitt, T., Filan, D., Daswani, M., & Hutter, M. (2016). Self-modification of policy and utility function in rational agents. In Artificial General Intelligence: 9th International Conference, AGI 2016, New York, NY, USA, July 16-19, 2016, Proceedings 9 (pp. 1-11). Springer International Publishing.

50. [**Fortune**] Fortune. (2023, July). Brainstorm Tech 2023: How Anthropic is paving the way for responsible A.I. [Video]. Fortune. https://fortune.com/videos/watch/brainstorm-tech-2023%3A-how-anthropic-is-paving-the-way-for-responsible-a.i./88517d5f-b5c3-40ac-b5bb-8368afc95acd

51. [**Field**] Field, S. (2025). Why do Experts Disagree on Existential Risk and P (doom)? A Survey of AI Experts. arXiv preprint arXiv:2502.14870.

52. [**Google**] Google DeepMind. (2024, April 9). Updating the Frontier Safety Framework. DeepMind. https://deepmind.google/discover/blog/updating-the-frontier-safety-framework/

53. [**Gryz**] Gryz, J., & Rojszczak, M. (2021). Black box algorithms and the rights of individuals: No easy solution to the "explainability" problem. Internet Policy Review, 10(2), 1-24.

54. [**IAISR**] Bengio, Y., Mindermann, S., Privitera, D., Besiroglu, T., Bommasani, R., Casper, S., ... & Zeng, Y. (2025). International AI Safety Report. arXiv preprint arXiv:2501.17805.

55. [**Gabriel**] Gabriel, I., Manzini, A., Keeling, G., Hendricks, L. A., Rieser, V., Iqbal, H., ... & Manyika, J. (2024). The ethics of advanced AI assistants. arXiv preprint arXiv:2404.16244.

56. [**GDM**] Shah, R., Irpan, A., Turner, A. M., Wang, A., Conmy, A., Lindner, D., ... & Dragan, A. (2025). An approach to technical AGI safety and security. Google DeepMind. https://storage.googleapis.com/deepmind-media/DeepMind.com/Blog/evaluating-potential-cybersecurity-threats-of-advanced-ai/An_Approach_to_Technical_AGI_Safety_Apr_2025.pdf

57. [**Greenblatt-A**] Greenblatt, R., Shlegeris, B., Sachan, K., & Roger, F. (2023). AI control: Improving safety despite intentional subversion. arXiv preprint arXiv:2312.06942.

58. [**Greenblatt-B**], R., Roger, F., Krasheninnikov, D., & Krueger, D. (2024). Stress-testing capability elicitation with password-locked models. arXiv preprint arXiv:2405.19550.

59. [**Griffin**] Griffin, C., Thomson, L., Shlegeris, B., & Abate, A. (2024). Games for AI control: Models of safety evaluations of AI deployment protocols. arXiv preprint arXiv:2409.07985.

60. [**Grosse**] Grosse, R., Bae, J., Anil, C., Elhage, N., Tamkin, A., Tajdini, A., ... & Bowman, S. R. (2023). Studying large language model generalization with influence functions. arXiv preprint arXiv:2308.03296.

61. [**Hadfield-Menell**] Hadfield-Menell, D., Russell, S. J., Abbeel, P., & Dragan, A. (2016). Cooperative inverse reinforcement learning. Advances in neural information processing systems, 29.

62. [**Hammond**] Hammond, L., Chan, A., Clifton, J., Hoelscher-Obermaier, J., Khan, A., McLean, E., ... & Rahwan, I. (2025). Multi-agent risks from advanced AI. arXiv preprint arXiv:2502.14143.

63. [**Hendrycks-A**] Hendrycks, D., Carlini, N., Schulman, J., & Steinhardt, J. (2021). Unsolved problems in ML safety. arXiv preprint arXiv:2109.13916.

64. [**Hendrycks-B**] Hendrycks, D. (2024). Systemic factors. In Introduction to AI safety, ethics, and society. Center for AI Safety. https://www.aisafetybook.com/textbook/systemic-factors

65. [**Hobbhahn**] Apollo Research. (2024, April 9). We need a science of evals. Apollo Research. https://www.apolloresearch.ai/blog/we-need-a-science-of-evals

66. [**Hofstätter**] Hofstätter, F., van der Weij, T., Teoh, J., Bartsch, H., & Ward, F. R. (2025). The Elicitation Game: Evaluating Capability Elicitation Techniques. arXiv preprint arXiv:2502.02180.

67. [**Huang**] Huang, T., Hu, S., Ilhan, F., Tekin, S. F., & Liu, L. (2024). Harmful fine-tuning attacks and defenses for large language models: A survey. arXiv preprint arXiv:2409.18169.

68. [**Hubinger-A**] Hubinger, E. (2020). An overview of 11 proposals for building safe advanced AI. arXiv preprint arXiv:2012.07532.

69. [**Hubinger-B**] Benton, J., Wagner, M., Christiansen, E., Anil, C., Perez, E., Srivastav, J., ... & Duvenaud, D. (2024). Sabotage evaluations for frontier models. arXiv preprint arXiv:2410.21514.

70. [**Hubinger-C**] Denison, C., MacDiarmid, M., Barez, F., Duvenaud, D., Kravec, S., Marks, S., ... & Hubinger, E. (2024). Sycophancy to subterfuge: Investigating reward-tampering in large language models. arXiv preprint arXiv:2406.10162.

71. [**Hubinger-D**] Hubinger, E., Denison, C., Mu, J., Lambert, M., Tong, M., MacDiarmid, M., ... & Perez, E. (2024). Sleeper agents: Training deceptive LLMs that persist through safety training. arXiv preprint arXiv:2401.05566.

72. [**Ilyas**] Engstrom, L., Feldmann, A., & Madry, A. (2024). DsDm: Model-aware dataset selection with datamodels. *arXiv preprint arXiv:2401.12926*.

73. [**International AI Safety Report-A**] UK Government. (2025). *International AI Safety Report 2025* (Accessible version). UK Department for Science, Innovation and Technology. https://assets.publishing.service.gov.uk/media/679a0c48a77d250007d313ee/International_AI_Safety_Report_2025_accessible_f.pdf

74. [**International AI Safety Report-B**] Bengio, Y., Mindermann, S., Privitera, D., Besiroglu, T., Bommasani, R., Casper, S., ... & Zeng, Y. (2025). International AI Safety Report. *arXiv preprint arXiv:2501.17805*.

75. [**Irving**] Irving, G., Christiano, P., & Amodei, D. (2018). AI safety via debate. arXiv preprint arXiv:1805.00899.

76. [**Ji**] Ji, J., Qiu, T., Chen, B., Zhang, B., Lou, H., Wang, K., ... & Gao, W. (2023). AI alignment: A comprehensive survey. arXiv preprint arXiv:2310.19852.

77. [**Jin**] Jin, H., Hu, L., Li, X., Zhang, P., Chen, C., Zhuang, J., & Wang, H. (2024). Jailbreakzoo: Survey, landscapes, and horizons in jailbreaking large language and vision-language models. arXiv preprint arXiv:2407.01599.

78. [**Jones**] Jones, E., Dragan, A., & Steinhardt, J. (2024). Adversaries can misuse combinations of safe models. arXiv preprint arXiv:2406.14595.

79. [**Jumper**] Jumper, J., Evans, R., Pritzel, A., Green, T., Figurnov, M., Ronneberger, O., ... & Hassabis, D. (2021). Highly accurate protein structure prediction with AlphaFold. nature, 596(7873), 583-589.

80. [**Khan**] Khan, A., Casper, S., & Hadfield-Menell, D. (2025). Randomness, Not Representation: The Unreliability of Evaluating Cultural Alignment in LLMs. arXiv preprint arXiv:2503.08688.

81. [**Keep the Future Human**] Aguirre, A. (2025). Keep the future human. Future of Life Institute. https://keep-thefuturehuman.ai/

82. [**Kirchenbauer**] Kirchenbauer, J., Geiping, J., Wen, Y., Katz, J., Miers, I., & Goldstein, T. (2023, July). A watermark for large language models. In International Conference on Machine Learning (pp. 17061-17084). PMLR.

83. [**Korbak**] Korbak, T., Balesni, M., Shlegeris, B., & Irving, G. (2025). How to evaluate control measures for LLM agents? A trajectory from today to superintelligence. arXiv preprint arXiv:2504.05259.

84. [**Ladish**] Bondarenko, A., Volk, D., Volkov, D., & Ladish, J. (2025). Demonstrating specification gaming in reasoning models. arXiv preprint arXiv:2502.13295.

85. [**Lawrence**] Lawrence, M., Shipman, M., Janzwood, S., Arnscheidt, C., Donges, J. F., Homer-Dixon, T., ... & Wunderling, N. (2024). Polycrisis Research and Action Roadmap-Gaps, opportunities, and priorities for polycrisis research and action.

86. [**Li-A**] Li, B., Qi, P., Liu, B., Di, S., Liu, J., Pei, J., ... & Zhou, B. (2023). Trustworthy AI: From principles to practices. ACM Computing Surveys, 55(9), 1-46.

87. [**Li-B**] Li, N., Pan, A., Gopal, A., Yue, S., Berrios, D., Gatti, A., ... & Hendrycks, D. (2024). The wmdp benchmark: Measuring and reducing malicious use with unlearning. arXiv preprint arXiv:2403.03218.

88. [**Longpre**] Longpre, S., Kapoor, S., Klyman, K., Ramaswami, A., Bommasani, R., Blili-Hamelin, B., ... & Henderson, P. (2024). A safe harbor for AI evaluation and red teaming. arXiv preprint arXiv:2403.04893.

89. [**MacDermott**] MacDermott, M., Fox, J., Belardinelli, F., & Everitt, T. (2024). Measuring Goal-Directedness. Advances in Neural Information Processing Systems, 37, 11412-11431.

90. [**Maini**] Maini, P., Goyal, S., Sam, D., Robey, A., Savani, Y., Jiang, Y., ... & Kolter, J. Z. (2025). Safety Pretraining: Toward the Next Generation of Safe AI. arXiv preprint arXiv:2504.16980.

91. [**Mallah**] Mallah, R. (2017). The landscape of AI safety and beneficence research: Input for brainstorming at Beneficial AI 2017. Future of Life Institute. https://futureoflife.org/landscape/ResearchLandscapeExtended.pdf

92. [**Marks**] Marks, S., Treutlein, J., Bricken, T., Lindsey, J., Marcus, J., Mishra-Sharma, S., ... & Hubinger, E. (2025). Auditing language models for hidden objectives. arXiv preprint arXiv:2503.10965.

93. [**Mazeika**] Mazeika, M., Yin, X., Tamirisa, R., Lim, J., Lee, B. W., Ren, R., Phan, L., Mu, N., Khoja, A., Zhang, O., & Hendrycks, D. (2025). Utility engineering: Analyzing and controlling emergent value systems in AIs. arXiv preprint arXiv:2502.08640. https://www.emergent-values.ai/

94. [**Michael**] Michael, J., Mahdi, S., Rein, D., Petty, J., Dirani, J., Padmakumar, V., & Bowman, S. R. (2023). Debate helps supervise unreliable experts. arXiv preprint arXiv:2311.08702.

95. [**Michaud**] Michaud, E. J., Liao, I., Lad, V., Liu, Z., Mudide, A., Loughridge, C., ... & Tegmark, M. (2024). Opening the AI Black Box: Distilling Machine-Learned Algorithms into Code. Entropy, 26(12), 1046, arXiv:2402.05110 (2024).

96. [**Murray**] Murray, M. (2025, April 9). AI risk management can learn a lot from other industries. AI Frontiers. https://www.ai-frontiers.org/articles/ai-risk-management-can-learn-a-lot-from-other-industries

97. [**Nevo**] Nevo, S., Lahav, D., Karput, A., Bar-On, Y., Bradley, H. A., & Alstott, J. (2024). Securing AI Model Weights. Technical report, RAND Corporation, 2024. https://www.rand.org/content/dam/rand/pubs/research_reports/RRA2800/RRA2849-1/RAND_RRA2849-1.pdf

98. [**Ngo**] Ngo, R., Chan, L., & Mindermann, S. (2022). The alignment problem from a deep learning perspective. arXiv preprint arXiv:2209.00626.

99. [**NIST**] National Institute of Standards and Technology. (2025, January 15). Updated guidelines for managing misuse risk for dual-use foundation models. https://www.nist.gov/news-events/news/2025/01/updated-guidelines-managing-misuse-risk-dual-use-foundation-models

100. [**Omohundro**] Omohundro, S. M. (2018). The basic AI drives. In Artificial intelligence safety and security (pp. 47-55). Chapman and Hall/CRC.

101. [**OpenAI-A**] OpenAI. (2023, December 18). Preparedness framework (beta). https://cdn.openai.com/openai-preparedness-framework-beta.pdf
102. [**OpenAI-B**] OpenAI. (2024). How we think about safety and alignment. https://openai.com/safety/how-we-think-about-safety-alignment/
103. [**OpenAI-C**] OpenAI. (2024, December 5). OpenAI o1 system card. https://cdn.openai.com/o1-system-card-20241205.pdf
104. [**OpenAI-D**] Baker, B., Huizinga, J., Gao, L., Dou, Z., Guan, M. Y., Madry, A., ... & Farhi, D. (2025). Monitoring reasoning models for misbehavior and the risks of promoting obfuscation. arXiv preprint arXiv:2503.11926.
105. [**OpenAI-E**] OpenAI. (2025, March 10). Detecting misbehavior in frontier reasoning models. https://openai.com/index/chain-of-thought-monitoring/
106. [**OpenAI-F**] Baker, B., Huizinga, J., Gao, L., Dou, Z., Guan, M. Y., Madry, A., ... & Farhi, D. (2025). Monitoring reasoning models for misbehavior and the risks of promoting obfuscation. arXiv preprint arXiv:2503.11926.
107. [**Paullada**] Paullada, A., Raji, I. D., Bender, E. M., Denton, E., & Hanna, A. (2021). Data and its (dis)contents: A survey of dataset development and use in machine learning research. Patterns, 2(11), 100336. https://doi.org/10.1016/j.patter.2021.100336
108. [**Phuong**] Phuong, M., Aitchison, M., Catt, E., Cogan, S., Kaskasoli, A., Krakovna, V., ... & Shevlane, T. (2024). Evaluating frontier models for dangerous capabilities. arXiv preprint arXiv:2403.13793.
109. [**Qi-A**] Qi, X., Zeng, Y., Xie, T., Chen, P. Y., Jia, R., Mittal, P., & Henderson, P. (2023). Fine-tuning aligned language models compromises safety, even when users do not intend to!. arXiv preprint arXiv:2310.03693.
110. [**Qi-B**] Qi, X., Wei, B., Carlini, N., Huang, Y., Xie, T., He, L., ... & Henderson, P. (2024). On evaluating the durability of safeguards for open-weight LLMs. arXiv preprint arXiv:2412.07097.
111. [**RAND**] Kulp, G., Gonzales, D., Smith, E., Heim, L., Puri, P., Vermeer, M. J. D., & Winkelman, Z. (2024). Hardware-enabled governance mechanisms: Developing technical solutions to exempt items otherwise classified under export control classification numbers 3A090 and 4A090 (RAND Working Paper WR-A3056-1). RAND Corporation. https://www.rand.org/content/dam/rand/pubs/working_papers/WRA3000/WRA3056-1/RAND_WRA3056-1.pdf
112. [**Raji**] Raji, I. D., Bender, E. M., Paullada, A., Denton, E., & Hanna, A. (2021). AI and the everything in the whole wide world benchmark. arXiv preprint arXiv:2111.15366.
113. [**Ren-A**] Ren, R., Basart, S., Khoja, A., Gatti, A., Phan, L., Yin, X., ... & Hendrycks, D. (2024). Safetywashing: Do AI Safety Benchmarks Actually Measure Safety Progress?. *Advances in Neural Information Processing Systems*, *37*, 68559-68594.
114. [**Ren-B**] Ren, R., Agarwal, A., Mazeika, M., Menghini, C., Vacareanu, R., Kenstler, B., ... & Hendrycks, D. (2025). The MASK Benchmark: Disentangling Honesty From Accuracy in AI Systems. *arXiv preprint arXiv:2503.03750*.
115. [**Reuel**] Reuel, A., Bucknall, B., Casper, S., Fist, T., Soder, L., Aarne, O., ... & Trager, R. (2024). Open problems in technical AI governance. arXiv preprint arXiv:2407.14981.
116. [**Russell**] Russell, S., Dewey, D., & Tegmark, M. (2015). Research priorities for robust and beneficial artificial intelligence. *AI magazine*, *36*(4), 105-114.
117. [**Sastry**] Sastry, G., Heim, L., Belfield, H., Anderljung, M., Brundage, M., Hazell, J., ... & Coyle, D. (2024). Computing power and the governance of artificial intelligence. *arXiv preprint arXiv:2402.08797*.
118. [**Savani**] Savani, Y., Trockman, A., Feng, Z., Schwarzschild, A., Robey, A., Finzi, M., & Kolter, J. Z. (2025). Antidistillation Sampling. arXiv preprint arXiv:2504.13146.
119. [**Scheurer**] Scheurer, J., Balesni, M., & Hobbhahn, M. (2023). Technical report: Large language models can strategically deceive their users when put under pressure. arXiv.
120. [**Shah-A**] Russell, S. (2020). *Human-compatible AI: A progress report*. Paper presented at the NeurIPS 2020 Workshop on Assistance. https://people.eecs.berkeley.edu/~russell/papers/neurips20ws-assistance
121. [**Shah-B**] Shah, R., Varma, V., Kumar, R., Phuong, M., Krakovna, V., Uesato, J., & Kenton, Z. (2022). Goal misgeneralization: Why correct specifications aren't enough for correct goals. arXiv preprint arXiv:2210.01790.
122. [**Sharkey**] Sharkey, L., Chughtai, B., Batson, J., Lindsey, J., Wu, J., Bushnaq, L., ... & McGrath, T. (2025). Open Problems in Mechanistic Interpretability. arXiv preprint arXiv:2501.16496.
123. [**Sharma**] Sharma, M., Tong, M., Korbak, T., Duvenaud, D., Askell, A., Bowman, S. R., ... & Perez, E. (2023). Towards understanding sycophancy in language models. arXiv preprint arXiv:2310.13548.
124. [**Shevlane**] Shevlane, T., Farquhar, S., Garfinkel, B., Phuong, M., Whittlestone, J., Leung, J., ... & Dafoe, A. (2023). Model evaluation for extreme risks. arXiv preprint arXiv:2305.15324.

125. [**Slattery**] Slater, P., Patel, A., & Rahimi, A. (2024). AI Risk Repository. Massachusetts Institute of Technology. https://airisk.mit.edu/
126. [**Soares**] Soares, N., Fallenstein, B., Yudkowsky, E., & Armstrong, S. (2015). Corrigibility. https://cdn.aaai.org/ocs/ws/ws0067/10124-45900-1-PB.pdf
127. [**Solaiman**] Solaiman, I., Talat, Z., Agnew, W., Ahmad, L., Baker, D., Blodgett, S. L., ... & Subramonian, A. (2023). Evaluating the social impact of generative AI systems in systems and society. arXiv preprint arXiv:2306.05949.
128. [**Sorensen**] Sorensen, T., Moore, J., Fisher, J., Gordon, M., Mireshghallah, N., Rytting, C. M., ... & Choi, Y. (2024). A roadmap to pluralistic alignment. arXiv preprint arXiv:2402.05070.
129. [**South**] South, T., Marro, S., Hardjono, T., Mahari, R., Whitney, C. D., Greenwood, D., ... & Pentland, A. (2025). Authenticated Delegation and Authorized AI Agents. arXiv preprint arXiv:2501.09674.
130. [**Thiel**] Thiel, D. (2023). Identifying and eliminating CSAM in generative ML training data and models. Stanford Internet Observatory, Cyber Policy Center, December, 23, 3.
131. [**Turpin**] Turpin, M., Michael, J., Perez, E., & Bowman, S. (2023). Language models don't always say what they think: Unfaithful explanations in chain-of-thought prompting. Advances in Neural Information Processing Systems, 36, 74952-74965.
132. [**UK AISI**] UK AI Security Institute. (2025). *AISI Challenge Fund: Priority research areas 2025.* https://cdn.prod.website-files.com/663bd486c5e4c81588db7a1d/67c99c8261da5261d5553893_AISI%20Challenge%20Fund_Priority%20Research%20Areas%202025%20(1).pdf
133. [**Wallach**] Wallach, H., Desai, M., Pangakis, N., Cooper, A. F., Wang, A., Barocas, S., ... & Jacobs, A. Z. (2024). Evaluating Generative AI Systems is a Social Science Measurement Challenge. arXiv preprint arXiv:2411.10939.
134. [**Wang**] Wang, S., Zhu, Y., Liu, H., Zheng, Z., Chen, C., & Li, J. (2024). Knowledge editing for large language models: A survey. ACM Computing Surveys, 57(3), 1-37.
135. [**Wasil-A**] Wasil, A., Smith, E., Katzke, C., & Bullock, J. (2024). AI Emergency Preparedness: Examining the federal government's ability to detect and respond to AI-related national security threats. arXiv preprint arXiv:2407.17347.
136. [**Wasil-B**] Wasil, A. (2024, June 7). What AI policy can learn from pandemic preparedness. Georgetown Security Studies Review. https://georgetownsecuritystudiesreview.org/2024/06/07/what-ai-policy-can-learn-from-pandemic-preparedness/
137. [**Wei**] Wei, A., Haghtalab, N., & Steinhardt, J. (2023). Jailbroken: How does LLM safety training fail?. Advances in Neural Information Processing Systems, 36, 80079-80110.
138. [**Weidinger-A**] Weidinger, L., Mellor, J., Rauh, M., Griffin, C., Uesato, J., Huang, P. S., ... & Gabriel, I. (2021). Ethical and social risks of harm from language models. arXiv preprint arXiv:2112.04359.
139. [**Weidinger-B**] Weidinger, L., Rauh, M., Marchal, N., Manzini, A., Hendricks, L. A., Mateos-Garcia, J., ... & Isaac, W. (2023). Sociotechnical safety evaluation of generative AI systems. arXiv preprint arXiv:2310.11986.
140. [**Weld**] Weld, D., & Etzioni, O. (2009). The First Law of Robotics: (A Call to Arms). In Safety and Security in Multi-agent Systems: Research Results from 2004-2006 (pp. 90-100). Berlin, Heidelberg: Springer Berlin Heidelberg.
141. [**Wen-A**] Wen, J., Zhong, R., Khan, A., Perez, E., Steinhardt, J., Huang, M., ... & Feng, S. (2024). Language models learn to mislead humans via RLHF. arXiv preprint arXiv:2409.12822.
142. [**Wen-B**] Wen, X., Lou, J., Lu, X., Yang, J., Liu, Y., Lu, Y., ... & Yu, X. (2025). Scalable Oversight for Superhuman AI via Recursive Self-Critiquing. arXiv preprint arXiv:2502.04675.
143. [**Yuan**] Yuan, Y., Jiao, W., Wang, W., Huang, J. T., He, P., Shi, S., & Tu, Z. (2023). GPT-4 is too smart to be safe: Stealthy chat with LLMs via cipher. *arXiv preprint arXiv:2308.06463.*
144. [**Zeng-A**] Zeng, Y., Lu, E. & Sun, K. (2025). Principles on symbiosis for natural life and living artificial intelligence. AI Ethics 5, 81–86. https://doi.org/10.1007/s43681-023-00364-8
145. [**Zeng-B**] Zhao, F., Wang, Y., Lu, E., Zhao, D., Han, B., Tong, H., ... & Zeng, Y. (2025). Redefining Superalignment: From Weak-to-Strong Alignment to Human-AI Co-Alignment to Sustainable Symbiotic Society. arXiv preprint arXiv:2504.17404.
146. [**Ziegler**] Ziegler, D., Nix, S., Chan, L., Bauman, T., Schmidt-Nielsen, P., Lin, T., ... & Thomas, N. (2022). Adversarial training for high-stakes reliability. Advances in neural information processing systems, 35, 9274-9286.
147. [**Zhou**] Zhou, Y., Liu, Y., Li, X., Jin, J., Qian, H., Liu, Z., ... & Yu, P. S. (2024). Trustworthiness in retrieval-augmented generation systems: A survey. arXiv preprint arXiv:2409.10102.
148. [**Zou**] Zou, A., Wang, Z., Carlini, N., Nasr, M., Kolter, J. Z., & Fredrikson, M. (2023). Universal and transferable adversarial attacks on aligned language models. arXiv preprint arXiv:2307.15043.

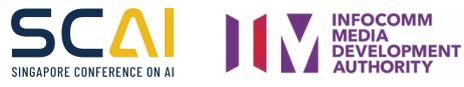

THE SINGAPORE CONSENSUS ON
GLOBAL AI SAFETY RESEARCH PRIORITIES

8 May 2025